\documentclass[11pt,a4paper]{article}
\usepackage{times,latexsym}

\usepackage{svg}
\usepackage{url}
\usepackage[T1]{fontenc}
\usepackage{graphicx}
\graphicspath{{./image/}}
\usepackage{enumitem}
\usepackage{booktabs}
\usepackage{amsmath}
\usepackage{longtable}
\usepackage{subcaption} 
\usepackage{caption}
\usepackage{array}
\usepackage{comment}
\usepackage{float}
\usepackage{pdflscape}
\usepackage{multirow}
\usepackage{lipsum} 
\usepackage{amssymb}
\usepackage{pifont}

\newcommand{\xmark}{\ding{55}}%
\usepackage[acceptedWithA]{tacl2021v1}
\usepackage{tacl2021v1}

\usepackage{xspace,mfirstuc,tabulary}

\newif\iftaclinstructions
\taclinstructionsfalse 
\iftaclinstructions
\renewcommand{\confidential}{}
\renewcommand{\anonsubtext}{(No author info supplied here, for consistency with
TACL-submission anonymization requirements)}
\newcommand{\instr}
\fi

\iftaclpubformat 

\else

\fi

\title{Direct Speech-to-Speech Neural Machine Translation: A Survey}

\author{
  Mahendra Gupta
  \\
  Dept. of CSE
  \\
  NITTTR Chandigarh, India
  \\
  \\
  \And
  Maitreyee Dutta
  \\
  Dept. of IMEE
  \\
  NITTTR Chandigarh, India
  \\
  \And
  Chandresh Kumar Maurya
  \\
  Dept. Of CSE
  \\
  IIT Indore, India
}

\date{}

\begin{document}
\maketitle
\begin{abstract}
Speech-to-Speech Translation (S2ST) models transform speech from one language to another target language with the same linguistic information. S2ST is important for bridging the communication gap among communities and has diverse applications. In recent years, researchers have introduced direct S2ST models, which have the potential to translate speech without relying on intermediate text generation, have better decoding latency, and the ability to preserve paralinguistic and non-linguistic features. However, direct S2ST has yet to achieve quality performance for seamless communication and still lags behind the cascade models in terms of performance, especially in real-world translation.  To the best of our knowledge, no comprehensive survey is available on the direct S2ST system, which beginners and advanced researchers can look upon for a quick survey. The present work provides a comprehensive review of direct S2ST models,  data and application issues, and performance metrics. We critically analyze the models’ performance over the benchmark datasets and provide research challenges and future directions.

\end{abstract}

\section{Introduction}
The S2ST task is the process of transforming speech in a source language to speech in a target language. It finds applications in live audio/video translation,  international conferences and meeting translations, educational video translation, and movie dubbing, to name a few. The rising computational power of computing devices, the expanding bandwidth of communication channels, and the progress in deep learning (DL) and machine learning (ML) techniques have enabled seamless communication.

Traditionally, the S2ST problem is solved by the cascade approach in that automatic speech recognition (ASR), machine translation (MT), and text-to-speech synthesis (TTS) models are glued together \cite{Laive_1997, Nakamura2006}. Another approach to building a cascade S2ST model is by chaining speech-to-text translation (ST) and TTS models. Nonetheless, the cascade models have been the de-facto choice for S2ST for a long time due to the easier availability of the pre-trained component models (ASR, MT, and TTS). However, they are being challenged by direct S2ST models in recent years owing to error propagation, higher training time, and memory cost of cascade models (refer to \S \ref{cascadevse2e} for more). 

Direct S2ST models translate source speech into target speech without using intermediate text representation. They are being popularised by the possibility of learning paralinguistic and non-linguistic features, including speaking style, emotions, energy, prosody, etc. \cite{Jia2019, Jia2021, Lee2022}. Languages without a writing system, often called unwritten languages, constitute 40\% of all languages \cite{lee-etal-2022-textless}. Text-based model training is not feasible for these languages.  The direct S2ST models can potentially address the challenge posed by unwritten languages \cite{Tjandra_2019_untranscribe, Zhang2021_UWSpeech, lee-etal-2022-textless}. Direct models follow an end-to-end (E2E) training approach, reducing error propagation and offering lower latency than cascade models \cite{Jia2019, Jia2021, Lee2022}. However, despite their advantages, these models encounter notable challenges such as: (a) getting sufficient parallel speech corpus in two different languages is extremely hard, thus hampering model development, (b) training with the speech of unwritten languages and their evaluation, (c) the potential threat of voice cloning, (d) absence of metrics directly taking generated and reference speech as input and returning a quality score, and (e) segmentation issue especially in simultaneous S2ST, and so on. Further advancements are still required for S2ST systems to attain the level of quality necessary for hassle-free communication.

\begin{figure*}
  \centering
  \begin{subfigure}[b]{0.55\textwidth}
    \centering
    \includegraphics[width=\textwidth]{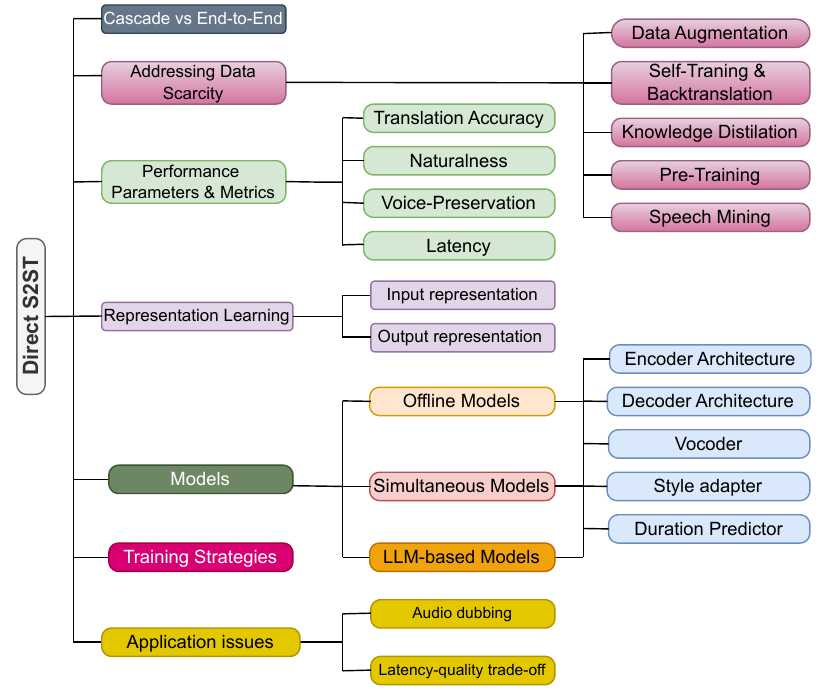}
    \caption{}
   \label{taxonomy}
  \end{subfigure}
  \begin{subfigure}[b]{0.4\textwidth}
    \centering
     \includegraphics[width=0.9\textwidth, height=8cm]
    {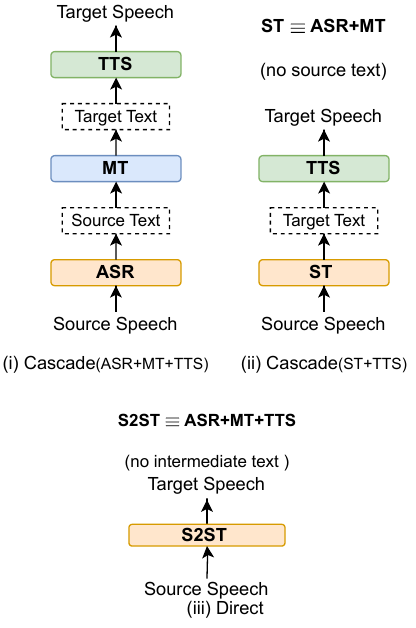}
    \caption{}
    \label{fig:S2ST_Cascade_Vs_Direct}
  \end{subfigure}
   \caption{(a) Direct S2ST Taxonomy (b) S2ST model design techniques (i) 3-stage cascade, (ii) 2-stage cascade, and (iii) direct.}
   \label{fig:taxonomy}
\end{figure*}

To provide an overview of research done in the S2ST field and a bucket of open problems,  the present work comprehensively reviews the direct S2ST literature, whose broad taxonomy is shown in Fig. \ref{taxonomy}.  Overall, the manuscript is organized as follows: \S\ref{Task_definition} defines the S2ST task
; \S\ref{cascadevse2e} presents the cascade vs E2E models; \S\ref{datapaucity} discusses the data scarcity. The performance metrics are presented in \S\ref{metrics} while \S\ref{repr} elucidates the segmentation and representation issues. Models and training strategies are discussed in \S\ref{sec:direct_s2st} and \S\ref{trainstr}, respectively. Application issues are discussed in \S\ref{application},  experiments in \S\ref{experiment}, challenges in \S\ref{challenges}, and finally concluded in \S\ref{challenges}.

\section{Task definition}\label{Task_definition}
Given a parallel speech corpus, denoted as $\mathcal{D}=\{(x_i,y_i)\}_{i=1}^n$. In this context, $x=\{f^s_1,\ldots, f^s_k\}$ and $y=\{f^t_1,\ldots, f^t_l\}$ represents the source and target speech utterances, respectively. Here, $f^s$ and $f^t$ refer to the source and target speech frames, and $k$ and $l$ represent utterance length in frames, respectively. The objective of direct S2ST models is to maximize the conditional probability $\mathcal{P}(y|x;\theta)$ as in \eqref{condobj} or minimize the negative log-likelihood loss \eqref{nll}.
\vspace{-0.3cm} 
\begin{equation} \label{condobj}
    \mathcal{P}(y|x;\theta)=\overset{k}{\underset{T=1}{\prod}}\mathcal{P}(f^t_T|f^t_{<T},x;\theta)
\end{equation}
\vspace{-0.5cm} 
\begin{equation} \label{nll}
\mathcal{L}_{(x,y)\in D}=-\overset{n}{\underset{i=1}{\sum}} \log \mathcal{P}(y_i|x_i;\theta)
\end{equation}
where $\mathcal{L}_{(x,y)\in D}$ is the cumulative loss on the dataset $D$ and $\theta$ is a model parameter.  Note that the problem formulation given in \eqref{condobj} is for \emph{Autoregressive} (AR) models \footnote{Non-autoregressive (NAR) models are an alternative modeling approach that have been proposed in the past few years.  Only a sparse number of works exist in the literature on S2ST where they use NAR in decoders of encoder-decoder frameworks. We discuss NAR briefly in \S \ref{decoderarch}}

On the other hand, cascade models have access to source transcripts denoted as a sequence of tokens (words) $w^s= \{w^s_1, \ldots, w^s_p\}$,  and the target transcript as a sequence of tokens, represented as  $w^t = \{w^t_1, \ldots, w^t_q\}$. In the 3-stage cascade model, the optimization objectives are as follows:
\vspace{-0.3cm} 
\begin{equation}
\small
    \mathcal{L}_{asr}=-\overset{n}{\underset{i=1}{\sum}} \log \mathcal{P}(w^s_i|x_i;\theta^{asr})
    \label{eq:asr_loss}
\end{equation}
\vspace{-0.3cm} 
\begin{equation}
\small
    \mathcal{L}_{mt}=-\overset{n}{\underset{i=1}{\sum}} \log \mathcal{P}(w^t_i|w^s_i;\theta^{mt})
    \label{eq:mt_loss}
\end{equation}
\vspace{-0.3cm} 
\begin{equation}
\small
    \mathcal{L}_{tts}=-\overset{n}{\underset{i=1}{\sum}} \log \mathcal{P}(f^t_i|w^t_i;\theta^{tts})
    \label{eq:tts_loss}
\end{equation}
\vspace{-0.3cm} 
\begin{equation}
\small
    \mathcal{L}_{st}=-\overset{n}{\underset{i=1}{\sum}} \log \mathcal{P}(w^t_i|x_i;\theta^{st})
    \label{eq:st_loss}
\end{equation}
 A 3-stage cascade model is built by independently minimizing the losses in \eqref{eq:asr_loss}, \eqref{eq:mt_loss}, and \eqref{eq:tts_loss}. A direct E2E ST model \cite{Berard2018_Audiobook, Kano2020} optimizes the loss in \eqref{eq:st_loss} (may also use losses in \eqref{eq:asr_loss}, \eqref{eq:mt_loss} in a multitask setup). This ST model is combined with the TTS model to form a 2-stage cascade S2ST model.

\section{Cascade vs. End-to-End S2ST models} \label{cascadevse2e}
The traditional S2ST systems follow a cascade architecture \cite{Laive_1997, Ney_1999_ST, Nakamura2006, Wahlster2000_Verbmobli}. They are designed either by chaining ASR, MT, and TTS or ST followed by TTS as illustrated in Figure \ref{fig:S2ST_Cascade_Vs_Direct} (i) and (ii) respectively. Either way, the cascade system relies on intermediate text. As such, they face several issues in modeling S2ST effectively. Firstly, they face challenges when dealing with low-resource languages lacking annotated corpora or unwritten languages \cite{Chen2022}. Secondly, paralinguistic features such as prosody, intonation, emotions, etc. are lost when representing speech via intermediate text, quintessential for building a real-world S2ST system. Thirdly, error propagation from one module to another \cite{Jia2019}, higher training time \cite{Huang2022_TranSpeech}, and memory footprint (due to 3 models vs 1 in direct S2ST) prohibit their application to low-powered devices. 

The above issues with cascade systems catapulted the development of direct S2ST systems bypassing the intermediate text generation, reducing training time and memory cost. There has been a lot of work developing direct S2ST models \cite{Jia2019, Jia2021, jia-etal-2022-cvss, Huang2022_TranSpeech, diwan2024textless}, etc. Therefore, it is imperative to compare cascade and direct models on quantitative and qualitative metrics (discussed in \S \ref{metrics}). Our literature survey reveals that there was a performance gap between cascade and direct models (both in terms of BLEU and mean opinion score (MOS)) \cite{Jia2019, Jia2021, lee-etal-2022-textless, zhu-etal-2023-diffs2ut, Huang2022_TranSpeech} which is now almost closed \cite{Chen2022, peng2024mslms2st}. These studies, however, are done on limited language pairs and may not generalize. Therefore, it remains to see an exhaustive comparison over multiple and distant language pairs involving large-scale datasets to truly establish the claim that the performance gap is indeed closed.

\section{Strategies for Addressing Data Scarcity}\label{datapaucity}
Parallel speech corpora are crucial for training direct S2ST models, and their scarcity can significantly affect their training, performance, and robustness. Creating parallel speech corpora requires substantial resource investment \cite{Chen2023_DataAugment_survey}, resulting in limited available datasets, as shown in Table \ref{Dataset_Stats} in the appendix. While this scarcity of supervised S2ST datasets prevents the application of direct S2ST systems to low-resource languages, techniques like data augmentation, pre-training, self-training, back-translation, knowledge distillation, and speech mining are employed to address data scarcity and enable S2ST for low-resource languages. These methods are further elaborated in the following sections.
\begin{figure*}
\centering
    \includegraphics[width=0.9\textwidth, height=3.4cm]{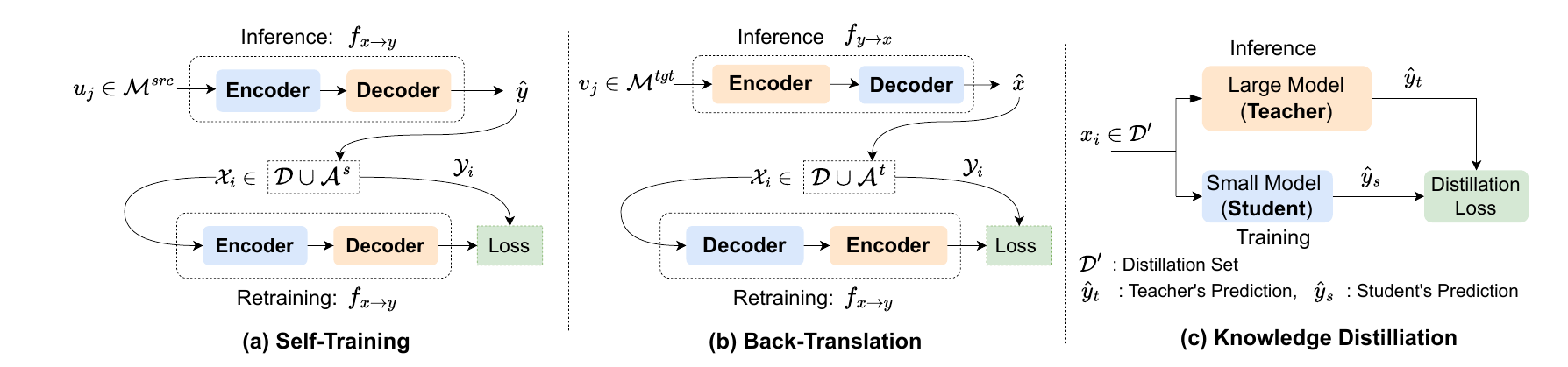}
    \captionsetup{font=small}
    \caption{Strategies for addressing data scarcity}
    \label{fig:BT_KD_ST}
\end{figure*}

\noindent \paragraph{Data Augmentation:} Data augmentation (DA) refers to techniques that boost the number of data points to enhance diversity without requiring the explicit collection of real supervised data points \cite{Feng2021_DataAug_survey}. The DA techniques suitable for various NLP tasks may not be directly applicable to S2ST since S2ST involves parallel speech, posing a challenge for augmenting existing data. Speech data can be augmented in various ways. For example, by adding
noise, speed and pitch perturbation, time and frequency masking, etc. 
In \cite{Jia2021}, data augmentation technique \emph{ConcatAug} is employed to preserve the speaker's voice on their turn. In particular,  training data is enhanced by randomly selecting pairs of training examples and concatenating their source speech, target speech, and target phoneme sequences to create new training examples. These new examples incorporate the voices of two different speakers in both the source and target speech, allowing the model to learn from instances that include speaker turns.
Many studies utilize synthesized speech from the target text in ST data \cite{Jia2019, Jia2021, Lee2022, nachmani_2023_translatotron3}, while others make use of MT and TTS models to label ASR corpora through pseudo-labeling \cite{Dong2022_Psudo_Labeled, Jia2022_Leveraging, Popuri2022_Enahancing_SelfSupervised,huang-etal-2023-xiaomi}.

\paragraph{Pre-Training:} Pre-training \cite{Popuri2022_Enahancing_SelfSupervised, Wei_2013_joint_pre-train_s2st} aims to improve the performance of the task in one domain with the learned knowledge from the tasks in another domain \cite{Pan2010_TransferLearning, Wang2022_Pre-Training_Application}. 
 Models learn salient features from an extensive training corpus during pre-training. Subsequently, the pre-trained model is applied to another (target) task, employing smaller datasets for fine-tuning~\cite{Van_dem_ordo_RepresentationLearning,tang-etal-2022-unified}. In recent works, self-supervised strategies are used for pre-training, to leverage the availability of large unlabeled datasets. Several pre-trained DL models, including BERT \cite{Devlin2019_BERT}, BART \cite{lewis-etal-2020-BART} (using unit mBART version), Wav2Vec 2.0 \cite{baevski2020_wav2vec2.0}, HuBERT \cite{Hsu_2021_HuBERT}, w2v-BERT \cite{Chung2021_w2V-BERT}, and VQ-VAE \cite{Van_Oord_2017_VQ-VAE}, which are trained on extensive unlabeled data, serve as the foundation for various NLP and speech tasks. The pre-trained models are used as encoders/decoders in direct S2ST models \cite{bansal2018pre, tsiamas-etal-2022-pretrained, Lee2022_Representation_Learning_for_Speech} (refer to \S \ref{sec:direct_s2st})
 
\paragraph{Self-training and Back-translation:} Self-training and Back-translation (BT) algorithms leverage monolingual data to train models requiring supervised data but lacking a supervised parallel corpus. These algorithms use the model's own confident predictions for monolingual data, which are initially trained on available parallel data \cite{edunov-etal-2018-understanding_BT, yang2022survey_semi_super}. Self-training utilizes source monolingual data, while BT utilizes target monolingual data to generate augmented data. Let us consider a utterance-wise parallel speech corpus denoted as $\mathcal{D}=\{(x_i, y_i)\}_{i=1}^n$, with $(x_i,y_i)$ representing the source and  target utterance, respectively. Additionally, we have source monolingual corpus $\mathcal{M}_{src} = \{u_j\}_{j = 1}^m$, and target monolingual corpus $\mathcal{M}_{tgt} = \{v_k\}_{k = 1}^q$, where \(m,q \gg n\). The forward translation model $f_{x\rightarrow y}$ is optimized for the parameter  \(\theta_{x\rightarrow y} = \text{argmin}_\theta \sum_{(x_i, y_i) \in \mathcal{D}} \mathcal{L}(f^\theta_{x\rightarrow y}(x_i), y_i)\), where $\mathcal{L}$ is the loss function. Similarly, another model in backward direction, $f_{y\rightarrow x}$, is optimized for the parameter \(\theta_{y\rightarrow x} = \text{argmin}_\theta \sum_{(x_i, y_i) \in \mathcal{D}} \mathcal{L}(f^\theta_{y\rightarrow x}(y_i), x_i)\). Using the already trained model \(f_{x \rightarrow y}\) and \(\mathcal{M}_{src}\), pseudo-labeled utterances \(\hat{y}\) are generated to create a new auxiliary parallel corpus $\mathcal{A}^s = \{(u_j, \hat{y}_j)\}_{j=1}^{m'}$
  by selecting confident predictions. In self-training, as shown in Figure \ref{fig:BT_KD_ST}(a), the model \(f_{x \rightarrow y}\) is retrained on the augmented data \(\mathcal{D} \cup \mathcal{A}^s\) \cite{He2020_revisiting_SelfTraining}. Similarly, an auxiliary parallel corpus \(\mathcal{A}^t = \{(\hat{x}_k, v_k)\}_{k=1}^{q'} \) is generated using the backward-trained model \(f_{y \rightarrow x}\). When employing back-translation (BT) the model \(f_{x \rightarrow y}\) is trained in the forward direction on the newly augmented data \(\mathcal{D} \cup \mathcal{A}^t\) as depicted in Figure \ref{fig:BT_KD_ST}(b) \cite{sennrich-etal-2016-improving}. Various studies utilize a denoising version of BT, where noise is added to the input \cite{fu_2023_improving_BackTranslation}. Several studies demonstrate that self-training \cite{Pino2020_ST_4ST} and BT \cite{Pino2020_ST_4ST} are highly beneficial, particularly for low-resource languages compared to high-resource ones.
These algorithms also benefit S2ST directly or indirectly, with the indirect method proving more effective \cite{Popuri2022_Enahancing_SelfSupervised,nachmani_2023_translatotron3}.

\paragraph{Knowledge Distillation:} Knowledge Distillation (KD) transfers learned knowledge from a large ensemble model to a smaller model as shown in Figure \ref{fig:BT_KD_ST}(c) \cite{Hinton2015_KnowledgeDistilation, Treviso2023_NLP_Methods_Survey}. The KD process is based on the teacher-student learning paradigm: the larger model serves as the teacher, while the smaller model acts as the student. 
KD proves to be valuable for tasks such as  ASR \cite{Hinton2015_KnowledgeDistilation}, ST \cite{Inaguma2021, Liu2019d_KnowledgeDistillation}, and S2ST \cite{Huang2022_TranSpeech} in scenarios with limited resources. In particular, AV-TranSpeech  \citet{Huang2022_TranSpeech} applies cross-modal distillation from audio-only pre-trained S2ST \cite{Popuri2022_Enahancing_SelfSupervised} to AV-TranSpeech. Doing so initializes the audio encoder and unit decoder and alleviates the low-resource problem of audio-visual data. Nonetheless, this approach remains relatively under-explored in the context of direct S2ST models.

\paragraph{Speech Mining:} Speech mining is an extension of \textbf{bitext mining} \cite{resnik-1998-parallel}, designed to discover parallel speech corpora from monolingual speech corpora. Bitext refers to text-to-text parallel data where the source and target sentences are in different languages. Bitext mining has been done in a supervised and unsupervised way (no cross-lingual resources like parallel text or bilingual lexicons are used) \cite{keung-etal-2020-unsupervised}. Multilingual fixed-length sentence embedding 
\cite{heffernan-etal-2022-bitext} using KD, contrastive learning \cite{tan2022bitext} are used for mining bitext in low-resource settings. Sentence embedding serves to cluster similar sentences closely together within the latent space \cite{schwenk-2018-filtering,artetxe-schwenk-2019-massively}. Similarly, for speech mining, multilingual fixed-length speech embedding is employed to represent the variable-length speech utterances. Speech mining has been used by a few works such as \citet{Duquenne_2021_Speech_Mining} where they utilize a teacher-student approach to train the multilingual speech embedding, which produces speech representations that are compatible with the output of text embedding layer.

\section{Performance Parameters \& Metrics} \label{metrics}
Offline S2ST systems are evaluated using Various text-based metrics such as BLEU 
 \cite{papineni-etal-2002-bleu}, ScareBLEU \cite{Post2018_scareBLEU}, BLEURT \cite{Sellam2020_BLEURT}, COMET \cite{Rei2020_COMET, Rei2022_COMET-22}, and chrF \cite{Popovic_2015_chrF} are employed to measure the quality of translated speech. These metrics are used after transcribing the speech through ASR due to the absence of direct evaluation metrics. These metrics depend on the performance of ASR models, and they also pose challenges for low-resource and unwritten languages. This is primarily due to either the unavailability of ASR systems or the existence of poor-quality models \cite{Salesky2021}. Recently, \citet{Chen2023_BLASER} introduced a metric called \textbf{BLASER}\footnote{Balanced Accuracy for Speech Evaluation by Recognition}, which is a text-free S2ST quality metric. BLASER takes source, reference, and generated speeches, embeds them to a common representation space via a shared encoder, and finally computes a score via a neural regressor. Formally, \emph{unsupervised} BLASER score is given below.
 \begin{equation}
     \text{BLASER}_U = \frac{cos(h_{src}, h_{mt})+cos(h_{mt},h_{ref})}{2}
 \end{equation}
 where cos(·, ·) is the cosine similarity function, $h_{src}, h_{mt}$, and $h_{ref}$ are the source, generated, and reference speech representations, respectively.  There is also a supervised BLASER score proposed \cite{Chen2023_BLASER}.
 
\textbf{Naturalness} measures how closely synthesized speech resembles natural speech. This is measured by a human-based subjective metric known as \textbf{MOS} (Mean Opinion Score). It is a numerical measure that represents the average subjective rating given by human listeners to the quality of a speech signal. MOS provides a simple and intuitive measure of speech quality that correlates well with human perception. Though MOS is widely used for speech quality, it has a few limitations such as subjective evaluation, bias, limited sample size,  time, and cost. Therefore, it is recommended also to report some \emph{objective} metrics along with MOS, especially those that capture the perception of speech quality. For example, the Perceptual Evaluation of Speech Quality (PESQ) \cite{rix2001perceptual}, its extension Perceptual Objective Listening Quality Analysis  (POLQA) \cite{beerends2013perceptual}, Short-Time Objective Intelligibility (STOI) \cite{taal2011algorithm} are a few objective algorithms designed to predict the perceived quality of speech as heard by human listeners.  

\textbf{Voice-preservation} measures the extent to which predicted speech is similar to a particular speaker's voice. It can be calculated using metrics for evaluating voice cloning. Measuring voice cloning involves a combination of subjective and objective methods to evaluate various aspects of the cloned voice, such as its similarity to the original voice (via speaker embedding similarity as done in \cite{Dong2023_PolyVoice}), naturalness, intelligibility, and acoustic properties. 

Simultaneous S2ST (Simul-S2ST) systems are evaluated using quality-based metrics for offline S2ST models along with \textbf{latency} metrics. Calculating the latency of the offline S2ST is straightforward: It is equal to the time elapsed in starting to produce the output from the decoder. For Simul-S2ST, calculating the average latency poses a significant challenge since it sees only the partial input. Due to the absence of such metrics, simultaneous text-to-text (Simul-T2T) latency metrics such as AP (average proportion) \cite{cho2016neural_AP} and AL (Average Lagging) \cite{ma-etal-2019-stacl_AL} are used as proxy. AL is further adapted for the Simul-S2T task \cite{Ren2020_SimulSpeech}, which is also used for the Simul-S2ST task with the help of ASR due to the non-availability of direct metrics. 

\textbf{Discussion:} Current practice in S2ST works is that authors report only text-based quality metrics ignoring objective and subjective metrics as described in this section. It is warranted that S2ST models are evaluated holistically using objective and subjective metrics as well. Further, the development of more effective textless S2ST quality metrics 
is recommended. 

\section{Segmentation \& Representation Learning} \label{repr}
S2ST models essentially take speech and optionally take text as inputs. Before we can train E2E S2ST models, we need to segment the speech and text followed by how to learn their representations. 
Handling long speech and text sequences is a challenging task \cite{kim2017joint, Tsiamas2022SHASAO}. The following section discusses segmentation and representation learning related to speech and text.
\subsection{Segmentation Learning}
Breaking down text into segments is a simpler task--it involves splitting based on robust punctuation marks used by current MT models. E2E S2ST models necessitate intricate speech segmentation, primarily because of the significant role played by the \emph{out-of-order} word relationships between input and output, alongside the absence of linguistic features. Traditionally, manual segmentation has been the norm for speech processing. However, due to its labor-intensive nature, there is a growing need for segmentation learning. Speech segmentation typically relies on either fixed-length splits which split the speech at fixed lengths, sometimes randomly \cite{Huang2022_TranSpeech} or \emph{pause} which splits the speech on Voice Activity Detection (VAD), as outlined by  \citet{Sohn1999ASM}. A third method, the \emph{hybrid} approach, integrates both length and linguistic cues for segmentation, as discussed by  \cite{Potapczyk2020SRPOLsSF, Gaido2021BeyondVA, Tsiamas2022SHASAO}. Notably, the hybrid approach demonstrates superior performance compared to length and pause-based methods \citep{Gaido2021BeyondVA}. However, there is still a gap in the hybrid and manual approaches to segmentation, and future work may consider paying attention to this.

\subsection{Representation Learning}
Representation learning is an important issue in S2ST because speech in two different languages are two different modalities that may reside in different representation spaces. Hence, we not only need better representation learning methods for speech and text but also their joint representation learning. 

Existing works leveraging text data for S2ST modeling utilize LSTM \cite{Kano2021_Transformer}, Transformers \cite{Lee2022}, etc. for representation learning. As the current trend is towards building \emph{textless} models, more recent works focus on learning efficient speech representations. Among them, the popular choices are \emph{raw waveform, spectrogram-based, unsupervised ASR}, and \emph{discrete-units based}. The raw waveform is the use of raw speech signal directly fed to the Sequence-to-Sequence (Seq2Seq) model and has been utilized by \cite{wang2023speechtospeech, KimLCL24}, inter alia. Mel-Filter Cepstral Coefficient (MFCC) feature, a spectrogram-based method, has been one of the most used speech representation methods \cite{Jia2019, jia2019leveraging_S2T, Tjandra_2019_untranscribe, Kano2021_Transformer, Lee2022, Huang2022_TranSpeech, Chen2023}, etc. where 80-dimension mel-spectrogram is computed.

Obtaining a substantial volume of labeled speech data for supervised feature representation learning poses significant challenges. Consequently, recent studies have turned to leveraging speech features acquired through \emph{unsupervised} and \emph{self-supervised}  methods. These approaches involve mapping continuous speech signals into discrete units-- akin to words and sub-words in the text domain. Such representations enable the integration of NLP tools into the speech domain. Among them, the popular choices are Wav2Vec \citep{schneider2019wav2vec} and its variants such as w2v-BERT \citep{Chung2021_w2V-BERT, Jia2022_Leveraging} and Wav2Vec 2.0 \citep{baevski2020_wav2vec2.0, Chen2022, song_2023_styles2st} and Hidden-Units BERT (HuBERT) \citep{hsu2021hubert}. What makes these representation methods take over the MFCC is that they can extract \emph{semantic-units}. Hence recent S2ST models invariably exploit HuBERT \cite{Huang2022_TranSpeech, wang2023speechtospeech, diwan2024textless, peng2024mslms2st, kaur2024direct} for semantic-unit discrete representation.  HuBERT utilizes a self-supervised approach to representation learning through masked prediction and employs $k$-means clustering to convert speech into discrete units \cite{Hsu_2021_HuBERT}.  Vector Quantized Variational Autoencoder (VQ-VAE) is another popular discrete unit representation model employing unsupervised speech representational learning \cite{Van_Oord_2017_VQ-VAE}. 
\begin{figure}
\centering
    \includegraphics[width=6cm, height=4cm]{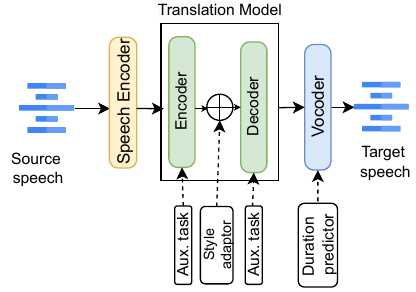}
    \captionsetup{font=small}
    \caption{Typical architecture of offline direct S2ST model}
    \label{fig:GenericDirectS2ST}
\end{figure}
\section{Direct S2ST Models}
\label{sec:direct_s2st}
Direct S2ST models can be classified broadly into three main categories: offline, simultaneous (Simul), and LLM-based S2ST\footnote{LLM-based S2ST is a very recent approach to S2ST modeling and is given special treatment due to their superior performance.}. Offline models, as the name suggests, start decoding after having seen the entire utterance, whereas simultaneous (aka streaming models) models can start decoding only with partial utterances. As such, Simul-S2ST is amenable to real-time translation, dubbing, etc. In this section, we will delve deeper into these three category models.  In particular, we will discuss the architectural level changes made in vanilla Seq2Seq to devise the offline/Simul/LLM-S2ST models. 
\subsection{Offline S2ST models}
As mentioned previously, offline direct S2ST models start decoding after having seen the entire utterance. The typical architecture of such models is shown in Fig. \ref{fig:GenericDirectS2ST} which is based on Seq2Seq with attention (not shown in Fig.). These models have \emph{translation model} component besides speech encoder (as mentioned in \S \ref{repr}) and vocoder (for speech synthesis). Dashed arrow components are optional and used in a few existing works. This section will discuss various design choices used by existing works for translation model component as well as vocoder. Style adapters and duration predictors will also be explained as they play an important role in designing offline direct S2ST models.
\subsubsection{Encoder Architecture}
Some models use separate \emph{speech encoders} to convert raw speech into spectrograms or discrete units, which are then fed into the translation model's encoder \cite{Jia2019, diwan2024textless}. Others integrate both into a single and directly feed raw speech into the translation model's encoder \cite{Popuri2022_Enahancing_SelfSupervised, Li2023_TextLess_Direct}.
For example, Translatotron 1 \cite{Jia2019} uses a spectrogram as input fed to the LSTM encoder. Conformer \cite{Gulati2020_Conformer} is used as the encoder in Translatotron 2 and 3 \cite{Jia2021, nachmani_2023_translatotron3}.
Essentially, recent direct S2ST models rely on transformers and their variants for encoder \cite{lee-etal-2022-textless, Li2023_TextLess_Direct}. To handle large speech inputs, there is often a downsampling module pre-pended before the encoder. 1-D CNN, Connectionist Temporal Classification (CTC) \cite{GravesConnectionistTC}, and adapters are commonly used for length mismatch \cite{lee-etal-2022-textless, Popuri2022_Enahancing_SelfSupervised}. As shown in Fig. \ref{fig:GenericDirectS2ST}, there is \emph{optional} auxiliary task module attached to the encoder. The auxiliary task again relies on transformer blocks to predict source and/or target phonemes/char/word or discrete units depending upon the text data availability \cite{Jia2021, lee-etal-2022-textless}. Dual-encoders disentangling semantic and acoustic information have been proposed \cite{le2024transvip}.

\subsubsection{Decoder Architecture} \label{decoderarch}
Similar to the encoder, the decoder of the translation model outputs either text \cite{Wang2022SimpleAE} or discrete unit \cite{lee-etal-2022-textless, duquenne_2022_SpeechMatrix, Li2023_TextLess_Direct}. Decoders in the existing works have been designed either in a \textbf{Autoregressive} (AR) or \textbf{Non-autoregressive} (NAR) way, as explained below. Further, they could be \textbf{single-pass} or \textbf{double-pass} decoders. Double-pass decoders employ a linguistic and an acoustic decoder, contrary to single-pass decoders with only an acoustic decoder. Due to multi-level speech processing and the ability to solve complex \emph{multimodal distribution}\footnote{S2ST is considered a multimodal problem due to (a) linguistic diversity during translation, and (b) diverse acoustic conditions, e.g., pitch, duration, energy, etc. }, double-pass decoders are shown to outperform single-pass decoders \cite{Jia2021, inaguma_2023_unity}.

In AR decoding, output tokens are predicted in a \emph{causal} manner, i.e., current output depends only on the previously generated tokens.   LSTMs and transformer-based decoders are preferred choices for autoregressive decoding. For example, unit mBART is used as a unit decoder in \cite{Popuri2022_Enahancing_SelfSupervised, diwan2024textless} whereas Translatotron 1 and 2 use  LSTMs. The CTC-based decoder is optionally added with the decoder as an auxiliary task \cite{lee-etal-2022-textless}. Despite high-quality translations, two-pass autoregressive decoders suffer from \emph{high-latency} issue because of their sequential generation. 

To address the high-latency decoding problem, NAR S2ST models have been proposed. NAR models can generate target speech in parallel by assuming conditional independence among the outputs. However, it becomes challenging to address the multimodal problem of the target speech via NAR compared to AR \cite{gu2018nonautoregressive}. 
To solve the linguistic modality problem, directed acyclic transformer \cite{pmlr-v162-huang22m}  and FastSpeech 2 as two-pass decoders are used by DASpeech \cite{Fang2023_DASpeech} whereas knowledge-distillation from AR model to NAR model is used in \cite{Huang2022_TranSpeech}. 

Besides AR and NAR decoding, there are other possible decoding choices. For example, it has been proven that translating multiple sentences of varying lengths can improve performance \cite{Huang2022_TranSpeech}. As such, \textbf{Top-K} length beam candidates are selected and decoded in parallel. \textbf{Noisy parallel decoding} is another option that uses AR decoding as a teacher to capture the more accurate optimum of the target distribution and compute the best translation for each fertility sequence \cite{Huang2022_TranSpeech}.

\subsubsection{Vocoder}
After the translation module is a vocoder whose job is to synthesize the speech from the decoder output which can be either text, discrete unit, or spectrogram. Recent and advanced speech models are adapted for synthesizing speech. For example, the majority of works leverage HiFi-GAN \cite{Kong2020_HiFiGAN} such as \cite{lee-etal-2022-textless,Popuri2022_Enahancing_SelfSupervised, Dong2022_Psudo_Labeled, diwan2024textless}, etc. as vocoder. Other speech synthesis models include non-autoregressive Transformer stack initialized by VQ-VAE \cite{Li2023_TextLess_Direct},  Light Convolution Blocks \cite{zhang_2022_direct_botelneck}, Non-Attentive Tacotron (NAT) \cite{shen2021nonattentive} which has duration prediction module, an LSTM module, and a residual convolutional block \cite{Jia2021}. Traditional estimation algorithms such as  Griffin  Lim \cite{GriffinLim} may also be used to convert the mel-spectrogram to waveform. To support multilingual S2ST, multilingual vocoders have been trained with language embedding and language ID as an auxiliary loss \cite{gong2023multilingual_S2ST}. 
\subsubsection{Style Adapter}
One of the desired characteristics of S2ST systems is that the translated speech should preserve (if need be) the original speaker's paralinguistic features, such as speaking rate, tone, intonation, etc. To this end, existing works insert \emph{style adapter} layer between the translator's encoder and decoder. \emph{Speech normalization} \cite{lee-etal-2022-textless} is one such technique to adapt to the source speaker's voice. The key idea is to extract discrete units from the reference speaker and then train the speech normalization module (which consists of pre-trained HuBERT  + CTC) using multi-speaker data as input and reference discrete units as targets. For example, Translatotron 1 uses a discriminatively pre-trained encoder on speaker verification to condition the decoder for voice preservation. Due to \textbf{voice cloning} issue, the aforementioned technique is unsuitable for production deployment. An approach to address this issue is to train on the same speaker's speech on the source and target side so that the model can transfer prosodic information at inference time \cite{Jia2021}. A side benefit that comes for free is that the S2ST model does not require speaker segmentation for translating multi-speaker utterances.
Interestingly, recent S2ST models implicitly model the para-/non-linguistic characteristics during decoding  \cite{Dong2023_PolyVoice, peng2024mslms2st}.

\subsubsection{Duration Predictor}
Duration predictor is a module often applied (but not always) along with the speech synthesizer. Its job is to predict the duration of each output element-- be it a phoneme or discrete unit, which is later converted to a waveform. For example, Translatotron 2 uses NAT TTS \cite{shen2021nonattentive}, which predicts the duration of each element followed by upsampling. However, unlike NAT, it optimizes $\mathcal{L}^2$ loss on the total duration predicted instead of \emph{per-phoneme} level which is costly to obtain.  TTS models like Fastspeech 2 \cite{Ren2020_FastSpeech2} have an in-built duration predictor, which has been used by \cite{Lee2022}. Works employing unit-to-speech conversion \cite{lee-etal-2022-textless, inaguma_2023_unity, Fang2023_DASpeech, zhu-etal-2023-diffs2ut, Shi_2023_Multiple_TTS} via HiFi-GAN TTS enhance the latter via duration prediction module from \cite{Ren2020_FastSpeech2}.
\subsection{Simultaneous (Simul-) S2ST Models}
 The models following the offline setting consider all the input speech to be available before commencing the translation process. In contrast, simultaneous models initiate translation upon receiving partial input speech \cite{agrawal-etal-IWSLT-2023-findings}. Simul-S2ST models may borrow encoders, decoders, and vocoders used by offline models (optionally style adapter and duration predictor as well) as mentioned in the previous section. However, they differ in how they process and segment the speech. 

Simul-S2ST is an important problem; however, simultaneously enhancing \textbf{ translation quality} while reducing \textbf{ latency} presents a formidable challenge.  The Simul-S2ST problem faces several issues in practical implementation; {\bf reordering,  acoustic ambiguity, variable speech rate, and long inputs} being prominent among them. 
One of the most important decisions in Simul-S2ST is to decide when to start the translation while balancing latency and quality. As such, there are \textbf{fixed} (like wait-$k$ \cite{ma-etal-2020-simulmt}) and \textbf{adaptive} policies ( like MILk, MMA \cite{Arivazhagan2019_MonotonicLookback, ma2022directMonotonicMultiHead}, etc. For more details on these policies, refer to \cite{Ma2019MonotonicMA}) proposed in Simul-MT and Simul-ST literature may be borrowed while designing effective and practical Simul-S2ST models. 

Traditional Simul-S2ST models (cascaded Simul-S2ST) have studied latency-quality trade-off under various policies \cite{zheng-etal-2020-fluent, dugan2023_when_to_speak} and find that no-single best-policy exist for all languages. Hence, it is recommended to tune the policy per language basis. The research on designing direct Simul-S2ST is severely limited. For example, a simultaneous policy, V-MMA (Variational Monotonic MultiHead Attention \cite{ma2022directMonotonicMultiHead}) considers every attention head as independent monotonic attention and models the alignment between the source and target sequences using latent variables instead of recurrent estimation which leads to an inferior estimate of the alignment matrix.  Direct Simul-S2ST using V-MMA policy, which adopts a discrete unit-based direct approach, reduces the average latency but compromises translation accuracy \cite{ma2022directMonotonicMultiHead}.

\subsection{Large Language Models (LLM)-S2ST}
Our third category of S2ST models is LLM-based. The recent success of  Generative Pre-Trained Transformers (GPT) \cite{ Openai_2018_GPT, Brown2020_GPT3,ouyang2022training_GPT} and BERT models \cite{Devlin2019_BERT} over various NLP tasks gives rise to what we know as LLMs. These models exhibit in-context learning (ICL) when trained on vast datasets. Extensive training unlocks their latent \emph{emergent abilities} \citep{Wei2022EmergentAO}, enabling them to perform few-shot and zero-shot learning through prompting. It alludes to using LLMs for S2ST tasks as well.  On a high level, LLM-based S2ST works leverage speech LM, which is prompted for speech generation in a target language. One issue is that LLMs operate on discrete tokens, whereas speech involves continuous values and cannot be directly applied to these models. Several research \cite{Wu2023_SpeechGen, zhang2023speak_forign_Languages, Dong2023_PolyVoice} utilize discrete-unit speech representation of the source/target speech. 

The effectiveness of LLM-based S2ST lies in several strategies, such as (a) what prompt to use? (b) how to do prompt-tuning? (c) which LM to use?  Works such as \cite{Wu2023_SpeechGen} use task ID as a prompt while others \cite{ zhang2023speak_forign_Languages, peng2024mslms2st, Dong2023_PolyVoice,  gong2024seamlessexpressivelm} use source and target semantic units and source acoustic units for prompting. Deep prompt tuning \cite{li-liang-2021-prefix}, chain-of-thought prompting \cite{gong2024seamlessexpressivelm}
have been explored in recent works. For LM, mBART \cite{Wu2023_SpeechGen}, VALL-E and its extension VALL-EX
\cite{zhang2023speak_forign_Languages} have been the preferred choice. Expressive S2ST \cite{gong2024seamlessexpressivelm} claims to preserve the speaker style without relying on aligned
speech-text data or speaker-aligned speech.  There are still some questions that need to be answered. For example, what is the best strategy for prompt design, and how to pre-train/fine-tune them parameter-efficiently for S2ST tasks? Further, the use of LLMs for Simul-MT has been recently proposed \citep{Agostinelli2023SimulLLMAF} and it remains to see how to adapt Simul-MT to Simul-S2ST.
\vspace{-.5pt}
\section{Training Strategies} \label{trainstr}
Training of E2E S2ST, in general, follows the training of DL models \cite{DBLP:journals/corr/abs-1206-5533}. Pre-training, self-supervised, unsupervised, and weakly supervised training approaches are primarily used to solve data scarcity as mentioned in \S \ref{datapaucity}. Therefore, we split our discussion on training based on the availability of \emph{external data}.
\vspace{-.5pt}
\subsection{Training with External Data}
Training of direct S2ST models optimizes the negative conditional log-likelihood as given in \eqref{nll}. However, there are cases when external sources and/or target transcripts are available along with target speech. Therefore, a natural question arises on how to leverage such external data. 
As depicted in Fig. \ref{fig:GenericDirectS2ST}, the architecture includes both an encoder and a decoder auxiliary task and is supervised by available labeled transcripts. Different sub-tasks are optimized simultaneously employing the E2E training approach \cite{Prabhavalkar2023_ASR_Survey, Tampuu2020_End-to-End_Driving}. Training with external data often invariably uses \textbf{Multitask Learning} (MTL) and is employed for \textbf{high-resource written} languages that have abundant text and/or speech data.

Several studies such as \citet{Jia2019, Jia2021, Kano2020_Transcoding, le2024transvip} employ MTL to propose the direct S2ST system. For example, Translatotron \cite{Jia2019} is the first direct S2ST model trained with two different setups: one with an MTL approach employing textual supervision using an auxiliary network of decoders that predict phonemes and the other without MTL. However, Translatotron's performance is significantly poor without MTL.

There are several issues present in Translatotron \cite{Jia2019} such as: (1) Auxiliary tasks for textual supervision are underutilized, as the learnings from auxiliary attention are not fully transferred to the main module, (2) the model faces difficulty mapping lengthy source speech spectrograms to target spectrograms, (3) over generation and under generation problem due to attention mechanism \cite{Ren2019_FastSpeech, Zheng_2019_E2E_TTS, Shen_2020_RobustControlled_TTS}. The existing bottlenecks are mitigated through architectural changes in Translatotron2 \cite{Jia2021} with four sub-modules. The single attention mechanism, based on Transformer \cite{Vaswani2017_Attention} alleviates the issue of lengthy spectrogram sequences by aligning them with shorter-length phonemes. Additionally, a Non-Attentive Tacotron (NAT) \cite{shen2021nonattentive} based TTS is employed to mitigate problems of over-generation and under-generation. Several models following the above architecture also use unsupervised data using techniques such as pre-training, BT, self-training, and pseudo-labeling to improve the performance of models \cite{Dong2022_Psudo_Labeled, Jia2022_Leveraging, nachmani_2023_translatotron3}. 


\subsection{Training w/o External Data aka Textless Training}
Compared to high-resource written languages, low-resource unwritten languages lack transcripts. For these languages, recent efforts in direct S2ST modeling propose \emph{textless} training. One approach is to train the model on speech spectrograms, but this method struggles to learn generalized patterns without text. Alternatively, textless training can use self-supervised (e.g., HuBERT) or unsupervised (e.g., VQ-VAE) discrete unit speech encoders.  This encoder converts continuous speech into discrete tokens (similar to text), enabling the application of textual NLP tools to speech.
For example, The textless model in \cite{Zhang2021_UWSpeech}  comprises three modules: converter, translator, and inverter. The converter transforms target speech into discrete units, the translator translates source speech into target discrete units, and the inverter inverts the predicted discrete units back into a speech waveform. Target discrete units enable the model to use the cross-entropy loss for model optimization. This architecture is beneficial for untranscribed languages \cite{Tjandra_2019_untranscribe, Zhang2021_UWSpeech,lee-etal-2022-textless, Huang2022_TranSpeech} or speech datasets without labelled transcripts \cite{Huang2022_TranSpeech, Lee2022}. The process of acquiring phonetic knowledge from languages that share syntactic similarities and have a written form might aid in learning representations for unwritten languages \cite{yallop2007_phonetics_phonology, Kuhl2008_phonetics_learning}. Nevertheless, the extent to which this assistance proves effective depends on the degree of similarity between the languages. Hence, leveraging the benefits of related languages with writing systems, XL-VAE \cite{Zhang2021_UWSpeech} is trained in a supervised manner by using phonemes from the related language as targets.

\section{Applications issues}\label{application}
Offline direct S2ST models find applications in the dubbing of movies and lectures. When deploying direct S2ST models for such applications, several things to be kept in mind. Firstly, getting a clean speech from a real-world environment is a challenging task. Cross-talk, noise, and background music removal must be done. Secondly, the challenge lies in handling various accents while translating, and voice preservation should be taken care of. The third important requirement of dubbing is {\bf isochrony}, i.e. dubbed speech has to closely match the timing of speech and pauses of the original audio. Inserting pauses, handling verbosity, and prosodic alignment are a few techniques to achieve isochrony. On the other hand, Simul-S2ST is more practical for real-world dubbing. Contrary to offline S2ST, these models require latency-quality handled in a better way along with isochrony.

\section{Experimental Results and Discussion}\label{experiment}
Table \ref{tab:performance_table} compares the performance of cascade and direct S2ST models from various studies. Results show a significant performance gap in BLEU scores between textless direct models without external data and cascade models (ID:1-4). The discrete unit-based Translatotron 2 \cite{Li2023_TextLess_Direct} outperforms other direct and cascade models in the Fisher Es$\rightarrow$En group (ID:5-10). \emph{UnitY} \cite{inaguma_2023_unity}, with dual decoders, surpasses all counterparts in its datasets and language directions and demonstrates performance comparable to cascade models (ID:14, 21, 26, 33). Utilizing text data through MTL and external data via pre-training benefits models \cite{Jia2019, Popuri2022_Enahancing_SelfSupervised, Lee2022}. Models with dual decoders outperform single-decoder models \cite{Jia2021, inaguma_2023_unity, Li2023_TextLess_Direct}. The MOS of direct models is slightly lower than cascade models perhaps due to training with synthesized target speech (ID:5-7,11,12), limiting their scores. MOS also depends on vocoder performance; lower translation quality negatively impacts MOS ratings \cite{Jia2019}. Using natural speech in the target can surpass the MOS of cascade models (ID:18, 24). Table \ref{tab:performance_Simul_LLM} shows results for Simultaneous and LLM-based models, though limited literature and no common comparison ground exist.

A standard dataset and language pair are necessary to compare the models fairly. Therefore, we implemented some existing models on the Es$\rightarrow$En language pair of the CVSS-C dataset, as shown in Table \ref{Experiment_result}, acknowledging that only some models are implemented due to code reproducibility issues and limited computing resources. The Fairseq library \cite{ott-etal-2019-fairseq} is used on a single NVIDIA Quadro GV100 GPU machine. The models (ID: 42-48) are trained from scratch with the provided hyperparameter settings. The results indicate that the performance of the direct model is deficient without MTL (ID: 45). However, when textual supervision is introduced through MTL, the performance of the discrete unit-based model \cite{Lee2022} shows significant improvement (ID: 46), while the spectrogram-based model \cite{Jia2019} still struggles with the provided amount of data (ID: 46). The model by \citet{Popuri2022_Enahancing_SelfSupervised}, which uses a pre-trained encoder and decoder and fine-tunes on CVSS-C dataset, outperforms all models trained without pre-training (ID: 49-52) and also outperforms the cascade 2-stage models, but falls short compared to the cascade 3-stage model. 



\section{Research Challenges and Future Directions}\label{challenges}
This section highlights challenges that may be explored by researchers working on S2ST problems. 

\subsection{Cascade vs End-to-End S2ST Models}
As discussed in \S \ref{cascadevse2e}, there has been limited empirical study comparing the cascade and E2E S2ST models. Furthermore, to our knowledge, no thorough assessment has been done for low-resource languages using E2E and cascade models. It may be interesting to compare E2E and cascade S2ST models on various S2ST datasets, particularly for low-resource and unwritten languages.
 
\subsection{S2ST on Code-Mix data}
Our literature review reveals a gap in research regarding the S2ST model applied to code-mix data. Code-mix data presents challenges like diverse lexicons, syntax variations, and a lack of labeled data. Hence, exploring the following questions would be intriguing: (a) Developing Code-mix S2ST datasets encompassing additional languages, (b) Evaluating the performance of current S2ST models on Code-mix S2ST data, and (c) Investigating whether pre-training across multiple languages aids in addressing code-mixing challenges. 

\subsection{Discrepancy between Automatic and Human Evaluation}
Current S2ST systems are evaluated by conducting ASR on the generated waveform and comparing it to the ground truth target text. Errors in ASR can impede evaluation via BLEU. As emphasized in \citep{marie-etal-2021-scientific}, the BLEU score is reported by more than 99\% of MT papers without considering statistical significance testing or human evaluation. Our examination of S2ST papers indicates a similar trend. The development of metrics that directly compare and evaluate source and target speech utterances without resorting to textual analysis remains an open challenge.

 \subsection{Multiple Speakers and Noise Handling }
In real-world scenarios, audio or video often feature multiple speakers in a noisy environment, each potentially with their accent, dialect, pitch, and tone. Conducting \emph{speech separation} beforehand could prove beneficial before inputting the data into the S2ST. Similarly, factors such as ambient noise, background music, cross-talk, and non-verbal sounds can pose challenges for S2ST model training. Distinguishing between meaningful speech and ambient noise presents a non-trivial task for the model.

\subsection{Multilingual and Simultaneous S2ST}
The recent surge in interest in multilingual S2ST is driven by its significance in real-world applications. For instance, a single speech may need to be delivered to multilingual communities, such as during a conference attended by a diverse audience. Multilingual S2ST can encompass various scenarios, including one-to-many, many-to-one, and many-to-many language translations. However, our literature review indicates a scarcity of research in this area. Moreover, there is an opportunity to explore simultaneous multilingual S2ST\footnote{In May 2024, Microsoft launched such a service in their Azure product.}, representing the most practical setting.

\subsection{Low-resource S2ST Datasets and Models}
Most current research efforts have concentrated on constructing S2ST models and datasets for languages with ample resources. However, given that the effectiveness of S2ST models hinges on parallel speech-to-speech corpora, there is a need for greater emphasis on developing datasets for low-resource languages. Thus, there is merit in constructing models that can transfer learning from language pairs with abundant resources to those with limited resources.

\subsection{Voice Cloning}
 The true challenge for direct models lies in attaining both lower latency and higher translation quality to fulfill real-time usage requirements. Simultaneously preserving the authenticity of the voice presents a challenge, as it is essential to guard against the misuse of voice cloning. 
 A hybrid model design combining discrete units and spectrograms can acquire linguistic and para-linguistic features, thereby enhancing translation quality, naturalness, and voice preservation. 

 \subsection{Faster Token Generation}
Many direct models primarily rely on autoregressive models as their sub-modules, in which the current output depends on previous inputs, resulting in a high degree of input dependency. Decreasing the number of modules or layers is essential to reduce the average latency of direct S2ST models. Moreover, shifting the focus from autoregressive models to non-autoregressive models is advisable. This shift in focus is crucial for enabling real-time usage scenarios.

\section{Conclusion}\label{concl}
We present a comprehensive survey of direct S2ST models. In particular, we discussed the comparison of E2E and cascade models, data scarcity, representation, and segmentation issues in S2ST modeling. A bird-eye view of architectures of offline, simultaneous, and LLM-based S2ST models is discussed in detail. We also offer training strategies and application issues that may be beneficial to industry practitioners. Finally, we enumerate a list of open problems and challenges that remain to be solved. Thus, we hope that the review will be a torch-bearer for someone starting their work in the S2ST domain.

\bibliography{tacl2021}

\begin{thebibliography}{138}
\expandafter\ifx\csname natexlab\endcsname\relax\def\natexlab#1{#1}\fi

\bibitem[{Agarwal et~al.(2023)Agarwal, Agrawal, Anastasopoulos, Bentivogli, Bojar, Borg, Carpuat, Cattoni, Cettolo, Chen, Chen, Choukri, Chronopoulou, Currey, Declerck, Dong, Duh, Est{\`e}ve, Federico, Gahbiche, Haddow, Hsu, Mon~Htut, Inaguma, Javorsk{\'y}, Judge, Kano, Ko, Kumar, Li, Ma, Mathur, Matusov, McNamee, P.~McCrae, Murray, Nadejde, Nakamura, Negri, Nguyen, Niehues, Niu, Kr.~Ojha, E.~Ortega, Pal, Pino, van~der Plas, Pol{\'a}k, Rippeth, Salesky, Shi, Sperber, St{\"u}ker, Sudoh, Tang, Thompson, Tran, Turchi, Waibel, Wang, Watanabe, and Zevallos}]{agrawal-etal-IWSLT-2023-findings}
Milind Agarwal, Sweta Agrawal, Antonios Anastasopoulos, Luisa Bentivogli, Ond{\v{r}}ej Bojar, Claudia Borg, Marine Carpuat, Roldano Cattoni, Mauro Cettolo, Mingda Chen, William Chen, Khalid Choukri, Alexandra Chronopoulou, Anna Currey, Thierry Declerck, Qianqian Dong, Kevin Duh, Yannick Est{\`e}ve, Marcello Federico, Souhir Gahbiche, Barry Haddow, Benjamin Hsu, Phu Mon~Htut, Hirofumi Inaguma, D{\'a}vid Javorsk{\'y}, John Judge, Yasumasa Kano, Tom Ko, Rishu Kumar, Pengwei Li, Xutai Ma, Prashant Mathur, Evgeny Matusov, Paul McNamee, John P.~McCrae, Kenton Murray, Maria Nadejde, Satoshi Nakamura, Matteo Negri, Ha~Nguyen, Jan Niehues, Xing Niu, Atul Kr.~Ojha, John E.~Ortega, Proyag Pal, Juan Pino, Lonneke van~der Plas, Peter Pol{\'a}k, Elijah Rippeth, Elizabeth Salesky, Jiatong Shi, Matthias Sperber, Sebastian St{\"u}ker, Katsuhito Sudoh, Yun Tang, Brian Thompson, Kevin Tran, Marco Turchi, Alex Waibel, Mingxuan Wang, Shinji Watanabe, and Rodolfo Zevallos. 2023.
\newblock \href {https://doi.org/10.18653/v1/2023.iwslt-1.1} {{FINDINGS} {OF} {THE} {IWSLT} 2023 {EVALUATION} {CAMPAIGN}}.
\newblock In \emph{Proceedings of the 20th International Conference on Spoken Language Translation (IWSLT 2023)}, pages 1--61, Toronto, Canada (in-person and online). Association for Computational Linguistics.

\bibitem[{Agostinelli et~al.(2023)Agostinelli, Wild, Raffel, Fuad, and Chen}]{Agostinelli2023SimulLLMAF}
Victor Agostinelli, Max Wild, Matthew Raffel, Kazi~Asif Fuad, and Lizhong Chen. 2023.
\newblock \href {https://arxiv.org/abs/2312.04691} {Simul-llm: A framework for exploring high-quality simultaneous translation with large language models}.
\newblock \emph{arXiv preprint arXiv:2312.04691}, abs/2312.04691.

\bibitem[{Agranovich et~al.(2024)Agranovich, Nachmani, Rybakov, Ding, Jia, Bar, Zen, and Ramanovich}]{agranovich2024_simultron}
Alex Agranovich, Eliya Nachmani, Oleg Rybakov, Yifan Ding, Ye~Jia, Nadav Bar, Heiga Zen, and Michelle~Tadmor Ramanovich. 2024.
\newblock \href {https://arxiv.org/pdf/2406.02133} {Simultron: On-device simultaneous speech to speech translation}.
\newblock \emph{arXiv preprint arXiv:2406.02133}.

\bibitem[{Arivazhagan et~al.(2019)Arivazhagan, Cherry, Macherey, Chiu, Google, Pang, Li, and Raffel}]{Arivazhagan2019_MonotonicLookback}
Naveen Arivazhagan, Colin Cherry, Wolfgang Macherey, Chung-Cheng Chiu, Semih~Yavuz Google, Ruoming Pang, Wei Li, and Colin Raffel. 2019.
\newblock \href {https://aclanthology.org/P19-1126.pdf} {Monotonic infinite lookback attention for simultaneous machine translation}.
\newblock pages 1313--1323, Florence, Italy. Association for Computational Linguistics.

\bibitem[{Artetxe and Schwenk(2019)}]{artetxe-schwenk-2019-massively}
Mikel Artetxe and Holger Schwenk. 2019.
\newblock \href {https://doi.org/10.1162/tacl_a_00288} {Massively multilingual sentence embeddings for zero-shot cross-lingual transfer and beyond}.
\newblock \emph{Transactions of the Association for Computational Linguistics}, 7:597--610.

\bibitem[{Baevski et~al.(2020)Baevski, Zhou, Mohamed, and Auli}]{baevski2020_wav2vec2.0}
Alexei Baevski, Yuhao Zhou, Abdelrahman Mohamed, and Michael Auli. 2020.
\newblock \href {https://proceedings.neurips.cc/paper_files/paper/2020/file/92d1e1eb1cd6f9fba3227870bb6d7f07-Paper.pdf} {wav2vec 2.0: A framework for self-supervised learning of speech representations}.
\newblock In \emph{Advances in Neural Information Processing Systems}, volume~33, pages 12449--12460, Vancouver, Canada. Curran Associates, Inc.

\bibitem[{Bansal et~al.(2019)Bansal, Kamper, Livescu, Lopez, and Goldwater}]{bansal2018pre}
Sameer Bansal, Herman Kamper, Karen Livescu, Adam Lopez, and Sharon Goldwater. 2019.
\newblock \href {https://doi.org/10.18653/v1/N19-1006} {Pre-training on high-resource speech recognition improves low-resource speech-to-text translation}.
\newblock In \emph{Proceedings of the 2019 Conference of the North {A}merican Chapter of the Association for Computational Linguistics: Human Language Technologies, Volume 1 (Long and Short Papers)}, pages 58--68, Minneapolis, Minnesota. Association for Computational Linguistics.

\bibitem[{Beerends et~al.(2013)Beerends, Schmidmer, Berger, Obermann, Pomy, and Keyhl}]{beerends2013perceptual}
John~G Beerends, Carsten Schmidmer, Jens Berger, Marc Obermann, J{\"u}rg Pomy, and Martin Keyhl. 2013.
\newblock \href {https://aes2.org/publications/elibrary-page/?id=16829} {Perceptual objective listening quality assessment (polqa), the third generation itu-t standard for end-to-end speech quality measurement part i—temporal alignment}.
\newblock \emph{Journal of the Audio Engineering Society}, 61(6):366--384.

\bibitem[{Bengio(2012)}]{DBLP:journals/corr/abs-1206-5533}
Yoshua Bengio. 2012.
\newblock \href {https://doi.org/10.1007/978-3-642-35289-8_26} {Practical recommendations for gradient-based training of deep architectures}.
\newblock In \emph{Neural Networks: Tricks of the Trade: Second Edition}, pages 437--478, Berlin, Heidelberg. Springer.

\bibitem[{Brown et~al.(2020)Brown, Mann, Ryder, Subbiah, Kaplan, Dhariwal, Neelakantan, Shyam, Sastry, Askell, Agarwal, Herbert-Voss, Krueger, Henighan, Child, Ramesh, Ziegler, Wu, Winter, Hesse, Chen, Sigler, Litwin, Gray, Chess, Clark, Berner, McCandlish, Radford, Sutskever, and Amodei}]{Brown2020_GPT3}
Tom Brown, Benjamin Mann, Nick Ryder, Melanie Subbiah, Jared~D Kaplan, Prafulla Dhariwal, Arvind Neelakantan, Pranav Shyam, Girish Sastry, Amanda Askell, Sandhini Agarwal, Ariel Herbert-Voss, Gretchen Krueger, Tom Henighan, Rewon Child, Aditya Ramesh, Daniel Ziegler, Jeffrey Wu, Clemens Winter, Chris Hesse, Mark Chen, Eric Sigler, Mateusz Litwin, Scott Gray, Benjamin Chess, Jack Clark, Christopher Berner, Sam McCandlish, Alec Radford, Ilya Sutskever, and Dario Amodei. 2020.
\newblock \href {https://proceedings.neurips.cc/paper_files/paper/2020/file/1457c0d6bfcb4967418bfb8ac142f64a-Paper.pdf} {Language models are few-shot learners}.
\newblock In \emph{Advances in Neural Information Processing Systems}, volume~33, pages 1877--1901. Curran Associates, Inc.

\bibitem[{Bérard et~al.(2018)Bérard, Besacier, Kocabiyikoglu, and Pietquin}]{Berard2018_Audiobook}
Alexandre Bérard, Laurent Besacier, Ali~Can Kocabiyikoglu, and Olivier Pietquin. 2018.
\newblock \href {https://doi.org/10.1109/ICASSP.2018.8461690} {End-to-end automatic speech translation of audiobooks}.
\newblock In \emph{2018 IEEE International Conference on Acoustics, Speech and Signal Processing (ICASSP)}, pages 6224--6228, Calgary, AB, Canada. IEEE.

\bibitem[{Chen et~al.(2023{\natexlab{a}})Chen, Tam, Raffel, Bansal, and Yang}]{Chen2023_DataAugment_survey}
Jiaao Chen, Derek Tam, Colin Raffel, Mohit Bansal, and Diyi Yang. 2023{\natexlab{a}}.
\newblock \href {https://doi.org/10.1162/TACL_A_00542/115238/AN-EMPIRICAL-SURVEY-OF-DATA-AUGMENTATION-FOR} {An empirical survey of data augmentation for limited data learning in nlp}.
\newblock \emph{Transactions of the Association for Computational Linguistics}, 11:191--211.

\bibitem[{Chen et~al.(2023{\natexlab{b}})Chen, Tam, Raffel, Bansal, and Yang}]{Chen2023}
Jiaao Chen, Derek Tam, Colin Raffel, Mohit Bansal, and Diyi Yang. 2023{\natexlab{b}}.
\newblock \href {https://doi.org/10.1162/TACL_A_00542} {An empirical survey of data augmentation for limited data learning in nlp}.
\newblock \emph{Transactions of the Association for Computational Linguistics}, 11:191--211.

\bibitem[{Chen et~al.(2023{\natexlab{c}})Chen, Duquenne, Andrews, Kao, Mourachko, Schwenk, and Costa-jussà}]{Chen2023_BLASER}
Mingda Chen, Paul-Ambroise Duquenne, Pierre Andrews, Justine Kao, Alexandre Mourachko, Holger Schwenk, and Marta~R. Costa-jussà. 2023{\natexlab{c}}.
\newblock \href {https://doi.org/10.18653/V1/2023.ACL-LONG.504} {Blaser: A text-free speech-to-speech translation evaluation metric}.
\newblock In \emph{Proceedings of the 61st Annual Meeting of the Association for Computational Linguistics}, volume~1, pages 9064--9079, Toronto, Canada. Association for Computational Linguistics.

\bibitem[{Chen et~al.(2023{\natexlab{d}})Chen, Tran, Yang, Du, Kao, Chung, Tomasello, Duquenne, Schwenk, Gong, Inaguma, Popuri, Wang, Pino, Hsu, and Lee}]{Chen2022}
Peng-Jen Chen, Kevin Tran, Yilin Yang, Jingfei Du, Justine Kao, Yu-An Chung, Paden Tomasello, Paul-Ambroise Duquenne, Holger Schwenk, Hongyu Gong, Hirofumi Inaguma, Sravya Popuri, Changhan Wang, Juan Pino, Wei-Ning Hsu, and Ann Lee. 2023{\natexlab{d}}.
\newblock \href {https://doi.org/10.18653/v1/2023.findings-acl.307} {Speech-to-speech translation for a real-world unwritten language}.
\newblock In \emph{Findings of the Association for Computational Linguistics: ACL 2023}, pages 4969--4983, Toronto, Canada. Association for Computational Linguistics.

\bibitem[{Cho and Esipova(2016)}]{cho2016neural_AP}
Kyunghyun Cho and Masha Esipova. 2016.
\newblock \href {https://doi.org/https://arxiv.org/pdf/1606.02012.pdf} {Can neural machine translation do simultaneous translation?}
\newblock \emph{arXiv preprint arXiv:1606.02012}.

\bibitem[{Chung et~al.(2021)Chung, Zhang, Han, Chiu, Qin, Pang, and Wu}]{Chung2021_w2V-BERT}
Yu-An Chung, Yu~Zhang, Wei Han, Chung-Cheng Chiu, James Qin, Ruoming Pang, and Yonghui Wu. 2021.
\newblock \href {https://doi.org/10.1109/ASRU51503.2021.9688253} {w2v-bert: Combining contrastive learning and masked language modeling for self-supervised speech pre-training}.
\newblock In \emph{2021 IEEE Automatic Speech Recognition and Understanding Workshop (ASRU)}, pages 244--250, Cartagena, Colombia. IEEE.

\bibitem[{Conneau et~al.(2022)Conneau, Ma, Khanuja, Zhang, Axelrod, Dalmia, Riesa, Rivera, and Bapna}]{Conneau2022_FLEURS}
Alexis Conneau, Min Ma, Simran Khanuja, Yu~Zhang, Vera Axelrod, Siddharth Dalmia, Jason Riesa, Clara Rivera, and Ankur Bapna. 2022.
\newblock \href {https://doi.org/10.1109/SLT54892.2023.10023141} {Fleurs: Few-shot learning evaluation of universal representations of speech}.
\newblock In \emph{2022 IEEE Spoken Language Technology Workshop, SLT 2022 - Proceedings}, pages 798--805. IEEE.

\bibitem[{Devlin et~al.(2019)Devlin, Chang, Lee, and Toutanova}]{Devlin2019_BERT}
Jacob Devlin, Ming~Wei Chang, Kenton Lee, and Kristina Toutanova. 2019.
\newblock \href {https://aclanthology.org/N19-1423.pdf} {Bert: Pre-training of deep bidirectional transformers for language understanding}.
\newblock In \emph{2019 Conference of the North American Chapter of the Association for Computational Linguistics: Human Language Technologies}, volume~1, pages 4171--4186, Minneapolis, Minnesota. Association for Computational Linguistics.

\bibitem[{Diwan et~al.(2024)Diwan, Srinivasan, Harwath, and Choi}]{diwan2024textless}
Anuj Diwan, Anirudh Srinivasan, David Harwath, and Eunsol Choi. 2024.
\newblock \href {http://arxiv.org/abs/2305.15405} {Textless low-resource speech-to-speech translation with unit language models}.
\newblock \emph{arXiv preprint arXiv:2305.15405}.

\bibitem[{Dong et~al.(2024)Dong, Huang, Tian, Xu, Ko, yunlong zhao, Feng, Li, Wang, Cheng, Yue, Bai, Chen, Lu, MA, Wang, Wang, and Wang}]{Dong2023_PolyVoice}
Qianqian Dong, Zhiying Huang, Qiao Tian, Chen Xu, Tom Ko, yunlong zhao, Siyuan Feng, Tang Li, Kexin Wang, Xuxin Cheng, Fengpeng Yue, Ye~Bai, Xi~Chen, Lu~Lu, Zejun MA, Yuping Wang, Mingxuan Wang, and Yuxuan Wang. 2024.
\newblock \href {https://openreview.net/forum?id=hCrFG9cyuC} {Polyvoice: Language models for speech to speech translation}.
\newblock In \emph{The Twelfth International Conference on Learning Representations}. OpenReview.net.

\bibitem[{Dong et~al.(2022)Dong, Yue, Ko, Wang, Bai, and Zhang}]{Dong2022_Psudo_Labeled}
Qianqian Dong, Fengpeng Yue, Tom Ko, Mingxuan Wang, Qibing Bai, and Yu~Zhang. 2022.
\newblock \href {https://doi.org/10.21437/Interspeech.2022-10011} {Leveraging pseudo-labeled data to improve direct speech-to-speech translation}.
\newblock In \emph{Proceedings of InterSpeech 2022}, Incheon, Korea. ISCA.

\bibitem[{Dugan et~al.(2023)Dugan, Wadhawan, Spence, Callison-Burch, McGuire, and Zordan}]{dugan2023_when_to_speak}
Liam Dugan, Anshul Wadhawan, Kyle Spence, Chris Callison-Burch, Morgan McGuire, and Victor~B. Zordan. 2023.
\newblock \href {https://www.isca-archive.org/interspeech_2023/dugan23_interspeech.html} {Learning when to speak: Latency and quality trade-offs for simultaneous speech-to-speech translation with offline models}.
\newblock In \emph{Proceedings of Interspeech 2023}, pages 5265--5266, Dublin, Ireland.

\bibitem[{Duquenne et~al.(2023)Duquenne, Gong, Dong, Du, Lee, Goswami, Wang, Pino, Sagot, and Schwenk}]{duquenne_2022_SpeechMatrix}
Paul-Ambroise Duquenne, Hongyu Gong, Ning Dong, Jingfei Du, Ann Lee, Vedanuj Goswami, Changhan Wang, Juan Pino, Beno{\^\i}t Sagot, and Holger Schwenk. 2023.
\newblock \href {https://doi.org/10.18653/v1/2023.acl-long.899} {{S}peech{M}atrix: A large-scale mined corpus of multilingual speech-to-speech translations}.
\newblock In \emph{Proceedings of the 61st Annual Meeting of the Association for Computational Linguistics}, pages 16251--16269, Toronto, Canada. Association for Computational Linguistics.

\bibitem[{Duquenne et~al.(2021)Duquenne, Gong, and Schwenk}]{Duquenne_2021_Speech_Mining}
Paul-Ambroise Duquenne, Hongyu Gong, and Holger Schwenk. 2021.
\newblock \href {https://proceedings.neurips.cc/paper/2021/hash/8466f9ace6a9acbe71f75762ffc890f1-Abstract.html} {Multimodal and multilingual embeddings for large-scale speech mining}.
\newblock In \emph{Advances in Neural Information Processing Systems}, volume~34, pages 15748--15761. Curran Associates, Inc.

\bibitem[{Edunov et~al.(2018)Edunov, Ott, Auli, and Grangier}]{edunov-etal-2018-understanding_BT}
Sergey Edunov, Myle Ott, Michael Auli, and David Grangier. 2018.
\newblock \href {https://doi.org/10.18653/v1/D18-1045} {Understanding back-translation at scale}.
\newblock In \emph{Proceedings of the 2018 Conference on Empirical Methods in Natural Language Processing}, pages 489--500, Brussels, Belgium. Association for Computational Linguistics.

\bibitem[{Fang et~al.(2023)Fang, Zhou, and Feng}]{Fang2023_DASpeech}
Qingkai Fang, Yan Zhou, and Yang Feng. 2023.
\newblock \href {https://github.com/ictnlp/DASpeech.} {Daspeech: Directed acyclic transformer for fast and high-quality speech-to-speech translation}.
\newblock \emph{Advances in Neural Information Processing Systems}, 36:72604--72623.

\bibitem[{Feng et~al.(2021)Feng, Gangal, Wei, Chandar, Vosoughi, Mitamura, and Hovy}]{Feng2021_DataAug_survey}
Steven~Y. Feng, Varun Gangal, Jason Wei, Sarath Chandar, Soroush Vosoughi, Teruko Mitamura, and Eduard Hovy. 2021.
\newblock \href {https://doi.org/10.18653/V1/2021.FINDINGS-ACL.84} {A survey of data augmentation approaches for nlp}.
\newblock In \emph{Findings of the Association for Computational Linguistics: ACL-IJCNLP 2021}, pages 968--988. Association for Computational Linguistics.

\bibitem[{Fu et~al.(2023)Fu, Tseng, Shi, Li, Hsu, Watanabe, and Lee}]{fu_2023_improving_BackTranslation}
Yu-Kuan Fu, Liang-Hsuan Tseng, Jiatong Shi, Chen-An Li, Tsu-Yuan Hsu, Shinji Watanabe, and Hung-yi Lee. 2023.
\newblock \href {https://arxiv.org/pdf/2305.07455} {Improving cascaded unsupervised speech translation with denoising back-translation}.
\newblock \emph{arXiv preprint arXiv:2305.07455}.

\bibitem[{Gaido et~al.(2021)Gaido, Negri, Cettolo, and Turchi}]{Gaido2021BeyondVA}
Marco Gaido, Matteo Negri, Mauro Cettolo, and Marco Turchi. 2021.
\newblock \href {https://aclanthology.org/2021.icnlsp-1.7} {Beyond voice activity detection: Hybrid audio segmentation for direct speech translation}.
\newblock In \emph{Proceedings of the 4th International Conference on Natural Language and Speech Processing (ICNLSP 2021)}, pages 55--62, Trento, Italy. Association for Computational Linguistics.

\bibitem[{Gong et~al.(2023)Gong, Dong, Popuri, Goswami, Lee, and Pino}]{gong2023multilingual_S2ST}
Hongyu Gong, Ning Dong, Sravya Popuri, Vedanuj Goswami, Ann Lee, and Juan Pino. 2023.
\newblock \href {https://arxiv.org/abs/2307.08655} {Multilingual speech-to-speech translation into multiple target languages}.
\newblock \emph{arXiv preprint arXiv:2307.08655}.

\bibitem[{Gong and Veluri(2024)}]{gong2024seamlessexpressivelm}
Hongyu Gong and Bandhav Veluri. 2024.
\newblock \href {https://arxiv.org/pdf/2405.20410} {Seamlessexpressivelm: Speech language model for expressive speech-to-speech translation with chain-of-thought}.
\newblock \emph{arXiv preprint arXiv:2405.20410}.

\bibitem[{Graves et~al.(2006)Graves, Fern\'{a}ndez, Gomez, and Schmidhuber}]{GravesConnectionistTC}
Alex Graves, Santiago Fern\'{a}ndez, Faustino Gomez, and J\"{u}rgen Schmidhuber. 2006.
\newblock \href {https://dl.acm.org/doi/abs/10.1145/1143844.1143891} {Connectionist temporal classification: labelling unsegmented sequence data with recurrent neural networks}.
\newblock In \emph{Proceedings of the 23rd International Conference on Machine Learning}, page 369–376, New York, NY, USA. Association for Computing Machinery.

\bibitem[{Griffin and Lim(1984)}]{GriffinLim}
D.~Griffin and Jae Lim. 1984.
\newblock \href {https://doi.org/10.1109/TASSP.1984.1164317} {Signal estimation from modified short-time fourier transform}.
\newblock \emph{IEEE Transactions on Acoustics, Speech, and Signal Processing}, 32(2):236--243.

\bibitem[{Gu et~al.(2018)Gu, Bradbury, Xiong, Li, and Socher}]{gu2018nonautoregressive}
Jiatao Gu, James Bradbury, Caiming Xiong, Victor~O.K. Li, and Richard Socher. 2018.
\newblock \href {https://openreview.net/forum?id=B1l8BtlCb} {Non-autoregressive neural machine translation}.
\newblock In \emph{6th International Conference on Learning Representations, {ICLR} 2018}, Vancouver, BC, Canada. OpenReview.net.

\bibitem[{Gulati et~al.(2020)Gulati, Qin, Chiu, Parmar, Zhang, Yu, Han, Wang, Zhang, Wu, and Pang}]{Gulati2020_Conformer}
Anmol Gulati, James Qin, Chung-Cheng Chiu, Niki Parmar, Yu~Zhang, Jiahui Yu, Wei Han, Shibo Wang, Zhengdong Zhang, Yonghui Wu, and Ruoming Pang. 2020.
\newblock \href {https://doi.org/10.21437/Interspeech.2020-3015} {{Conformer: Convolution-augmented Transformer for Speech Recognition}}.
\newblock In \emph{Proceedings of Interspeech 2020}, pages 5036--5040. ISCA.

\bibitem[{He et~al.(2020)He, Gu, Shen, and Ranzato}]{He2020_revisiting_SelfTraining}
Junxian He, Jiatao Gu, Jiajun Shen, and Marc'Aurelio Ranzato. 2020.
\newblock \href {https://openreview.net/forum?id=SJgdnAVKDH} {Revisiting self-training for neural sequence generation}.
\newblock In \emph{8th International Conference on Learning Representations, {ICLR} 2020, Addis Ababa, Ethiopia, April 26-30, 2020}, Addis Ababa, Ethiopia. OpenReview.net.

\bibitem[{Heffernan et~al.(2022)Heffernan, {\c{C}}elebi, and Schwenk}]{heffernan-etal-2022-bitext}
Kevin Heffernan, Onur {\c{C}}elebi, and Holger Schwenk. 2022.
\newblock \href {https://doi.org/10.18653/v1/2022.findings-emnlp.154} {Bitext mining using distilled sentence representations for low-resource languages}.
\newblock In \emph{Findings of the Association for Computational Linguistics: EMNLP 2022}, pages 2101--2112, Abu Dhabi, United Arab Emirates. Association for Computational Linguistics.

\bibitem[{Hinton et~al.(2015)Hinton, Vinyals, and Dean}]{Hinton2015_KnowledgeDistilation}
Geoffrey Hinton, Oriol Vinyals, and Jeff Dean. 2015.
\newblock \href {https://arxiv.org/abs/1503.02531v1} {Distilling the knowledge in a neural network}.
\newblock \emph{arXiv preprint arXiv:1503.02531}.

\bibitem[{Hsu et~al.(2021{\natexlab{a}})Hsu, Bolte, Tsai, Lakhotia, Salakhutdinov, and Mohamed}]{Hsu_2021_HuBERT}
Wei~Ning Hsu, Benjamin Bolte, Yao Hung~Hubert Tsai, Kushal Lakhotia, Ruslan Salakhutdinov, and Abdelrahman Mohamed. 2021{\natexlab{a}}.
\newblock \href {https://doi.org/10.1109/TASLP.2021.3122291} {Hubert: Self-supervised speech representation learning by masked prediction of hidden units}.
\newblock \emph{IEEE/ACM Transactions on Audio Speech and Language Processing}, 29:3451--3460.

\bibitem[{Hsu et~al.(2021{\natexlab{b}})Hsu, Bolte, Tsai, Lakhotia, Salakhutdinov, and Mohamed}]{hsu2021hubert}
Wei-Ning Hsu, Benjamin Bolte, Yao-Hung~Hubert Tsai, Kushal Lakhotia, Ruslan Salakhutdinov, and Abdelrahman Mohamed. 2021{\natexlab{b}}.
\newblock Hubert: Self-supervised speech representation learning by masked prediction of hidden units.
\newblock \emph{IEEE/ACM Transactions on Audio, Speech, and Language Processing}, 29:3451--3460.

\bibitem[{Huang et~al.(2022)Huang, Zhou, Liu, Li, and Huang}]{pmlr-v162-huang22m}
Fei Huang, Hao Zhou, Yang Liu, Hang Li, and Minlie Huang. 2022.
\newblock \href {https://proceedings.mlr.press/v162/huang22m.html} {Directed acyclic transformer for non-autoregressive machine translation}.
\newblock In \emph{Proceedings of the 39th International Conference on Machine Learning}, volume 162 of \emph{Proceedings of Machine Learning Research}, pages 9410--9428. PMLR.

\bibitem[{Huang et~al.(2023{\natexlab{a}})Huang, Liu, Liu, Ren, Zhang, He, and Zhao}]{Huang2022_TranSpeech}
Rongjie Huang, Jinglin Liu, Huadai Liu, Yi~Ren, Lichao Zhang, Jinzheng He, and Zhou Zhao. 2023{\natexlab{a}}.
\newblock \href {https://openreview.net/pdf?id=UVAmFAtC5ye} {Transpeech: Speech-to-speech translation with bilateral perturbation}.
\newblock In \emph{Proceedings of International Conference on Learning Representation (ICLR 2023)}, Kigali, Rwanda. openReview.net.

\bibitem[{Huang et~al.(2023{\natexlab{b}})Huang, Liu, Li, Tian, Yang, Zhang, Luan, Wang, Guo, and Su}]{huang-etal-2023-xiaomi}
Wuwei Huang, Mengge Liu, Xiang Li, Yanzhi Tian, Fengyu Yang, Wen Zhang, Jian Luan, Bin Wang, Yuhang Guo, and Jinsong Su. 2023{\natexlab{b}}.
\newblock \href {https://doi.org/10.18653/v1/2023.iwslt-1.39} {The xiaomi {AI} lab{'}s speech translation systems for {IWSLT} 2023 offline task, simultaneous task and speech-to-speech task}.
\newblock In \emph{Proceedings of the 20th International Conference on Spoken Language Translation (IWSLT 2023)}, pages 411--419, Toronto, Canada. Association for Computational Linguistics.

\bibitem[{Inaguma et~al.(2021)Inaguma, Kawahara, and Watanabe}]{Inaguma2021}
Hirofumi Inaguma, Tatsuya Kawahara, and Shinji Watanabe. 2021.
\newblock \href {https://doi.org/10.18653/v1/2021.naacl-main.150} {Source and target bidirectional knowledge distillation for end-to-end speech translation}.
\newblock In \emph{Proceedings of the 2021 Conference of the North American Chapter of the Association for Computational Linguistics: Human Language Technologies}, pages 1872--1881. Association for Computational Linguistics.

\bibitem[{Inaguma et~al.(2023)Inaguma, Popuri, Kulikov, Chen, Wang, Chung, Tang, Lee, Watanabe, and Pino}]{inaguma_2023_unity}
Hirofumi Inaguma, Sravya Popuri, Ilia Kulikov, Peng-Jen Chen, Changhan Wang, Yu-An Chung, Yun Tang, Ann Lee, Shinji Watanabe, and Juan Pino. 2023.
\newblock \href {https://doi.org/10.18653/v1/2023.acl-long.872} {{U}nit{Y}: Two-pass direct speech-to-speech translation with discrete units}.
\newblock In \emph{Proceedings of the 61st Annual Meeting of the Association for Computational Linguistics}, pages 15655--15680, Toronto, Canada. Association for Computational Linguistics.

\bibitem[{Jeuris and Niehues(2022)}]{jeuris-niehues-2022-libris2s}
Pedro Jeuris and Jan Niehues. 2022.
\newblock \href {https://aclanthology.org/2022.lrec-1.98} {{L}ibri{S}2{S}: A {G}erman-{E}nglish speech-to-speech translation corpus}.
\newblock In \emph{Proceedings of the Thirteenth Language Resources and Evaluation Conference}, pages 928--935, Marseille, France. European Language Resources Association.

\bibitem[{Jia et~al.(2022{\natexlab{a}})Jia, Ding, Bapna, Cherry, Zhang, Conneau, and Morioka}]{Jia2022_Leveraging}
Ye~Jia, Yifan Ding, Ankur Bapna, Colin Cherry, Yu~Zhang, Alexis Conneau, and Nobuyuki Morioka. 2022{\natexlab{a}}.
\newblock \href {https://doi.org/10.21437/Interspeech.2022-10938} {Leveraging unsupervised and weakly-supervised data to improve direct speech-to-speech translation}.
\newblock volume 2022-September, pages 1721--1725. International Speech Communication Association.

\bibitem[{Jia et~al.(2019{\natexlab{a}})Jia, Johnson, Macherey, Weiss, Cao, Chiu, Ari, Laurenzo, and Wu}]{jia2019leveraging_S2T}
Ye~Jia, Melvin Johnson, Wolfgang Macherey, Ron~J Weiss, Yuan Cao, Chung-Cheng Chiu, Naveen Ari, Stella Laurenzo, and Yonghui Wu. 2019{\natexlab{a}}.
\newblock \href {https://ieeexplore.ieee.org/stamp/stamp.jsp?arnumber=8683343} {Leveraging weakly supervised data to improve end-to-end speech-to-text translation}.
\newblock In \emph{IEEE International Conference on Acoustics, Speech and Signal Processing (ICASSP)}, pages 7180--7184. IEEE.

\bibitem[{Jia et~al.(2022{\natexlab{b}})Jia, Ramanovich, Remez, and Pomerantz}]{Jia2021}
Ye~Jia, Michelle~Tadmor Ramanovich, Tal Remez, and Roi Pomerantz. 2022{\natexlab{b}}.
\newblock \href {https://proceedings.mlr.press/v162/jia22b/jia22b.pdf} {Translatotron 2: High-quality direct speech-to-speech translation with voice preservation}.
\newblock In \emph{Proceedings of the 39 th International Conference on Machine Learning}, volume 162, pages 10120--10134, Baltimore, Maryland, USA. PMLR.

\bibitem[{Jia et~al.(2022{\natexlab{c}})Jia, Tadmor~Ramanovich, Wang, and Zen}]{jia-etal-2022-cvss}
Ye~Jia, Michelle Tadmor~Ramanovich, Quan Wang, and Heiga Zen. 2022{\natexlab{c}}.
\newblock \href {https://aclanthology.org/2022.lrec-1.720} {{CVSS} corpus and massively multilingual speech-to-speech translation}.
\newblock In \emph{Proceedings of the Thirteenth Language Resources and Evaluation Conference}, pages 6691--6703, Marseille, France. European Language Resources Association.

\bibitem[{Jia et~al.(2019{\natexlab{b}})Jia, Weiss, Biadsy, Macherey, Johnson, Chen, and Google}]{Jia2019}
Ye~Jia, Ron~J Weiss, Fadi Biadsy, Wolfgang Macherey, Melvin Johnson, Zhifeng Chen, and Yonghui~Wu Google. 2019{\natexlab{b}}.
\newblock \href {https://doi.org/10.21437/Interspeech.2019-1951} {Direct speech-to-speech translation with a sequence-to-sequence model}.
\newblock In \emph{Proceedings of INTERSPEECH 2019,}, pages 1123--1127, Graz, Austria. ISCA.

\bibitem[{Kano et~al.(2020{\natexlab{a}})Kano, Sakti, and Nakamura}]{Kano2020}
Takatomo Kano, Sakriani Sakti, and Satoshi Nakamura. 2020{\natexlab{a}}.
\newblock \href {https://doi.org/10.1109/TASLP.2020.2986886} {End-to-end speech translation with transcoding by multi-task learning for distant language pairs}.
\newblock \emph{IEEE/ACM Transactions on Audio Speech and Language Processing}, 28:1342--1355.

\bibitem[{Kano et~al.(2020{\natexlab{b}})Kano, Sakti, and Nakamura}]{Kano2020_Transcoding}
Takatomo Kano, Sakriani Sakti, and Satoshi Nakamura. 2020{\natexlab{b}}.
\newblock \href {https://doi.org/10.1109/TASLP.2020.2986886} {End-to-end speech translation with transcoding by multi-task learning for distant language pairs}.
\newblock \emph{IEEE/ACM Transactions on Audio, Speech, and Language Processing}, 28:1342--1355.

\bibitem[{Kano et~al.(2021)Kano, Sakti, and Nakamura}]{Kano2021_Transformer}
Takatomo Kano, Sakriani Sakti, and Satoshi Nakamura. 2021.
\newblock \href {https://doi.org/10.1109/SLT48900.2021.9383496} {Transformer-based direct speech-to-speech translation with transcoder}.
\newblock In \emph{2021 IEEE Spoken Language Technology Workshop (SLT)}, pages 958--965, Shenzhen, China. IEEE.

\bibitem[{Kaur et~al.(2024)Kaur, Bush, and Shi}]{kaur2024direct}
Prabhjot Kaur, L~Andrew~M Bush, and Weisong Shi. 2024.
\newblock \href {https://arxiv.org/abs/2402.15967} {Direct punjabi to english speech translation using discrete units}.
\newblock \emph{arXiv preprint arXiv:2402.15967}.

\bibitem[{Keung et~al.(2020)Keung, Salazar, Lu, and Smith}]{keung-etal-2020-unsupervised}
Phillip Keung, Julian Salazar, Yichao Lu, and Noah~A. Smith. 2020.
\newblock \href {https://doi.org/10.1162/tacl_a_00348} {Unsupervised bitext mining and translation via self-trained contextual embeddings}.
\newblock \emph{Transactions of the Association for Computational Linguistics}, 8:828--841.

\bibitem[{Kim et~al.(2024{\natexlab{a}})Kim, Lee, Choi, and Lee}]{KimLCL24}
Seung{-}Bin Kim, Sang{-}Hoon Lee, Ha{-}Yeong Choi, and Seong{-}Whan Lee. 2024{\natexlab{a}}.
\newblock \href {https://doi.org/10.1109/TASLP.2023.3349053} {Audio super-resolution with robust speech representation learning of masked autoencoder}.
\newblock \emph{{IEEE} {ACM} Transaction on Audio Speech and Language Processing}, 32:1012--1022.

\bibitem[{Kim et~al.(2024{\natexlab{b}})Kim, Lee, and Lee}]{Kim_2024_TranSentence}
Seung-Bin Kim, Sang-Hoon Lee, and Seong-Whan Lee. 2024{\natexlab{b}}.
\newblock \href {https://doi.org/10.1109/ICASSP48485.2024.10447331} {Transentence: speech-to-speech translation via language-agnostic sentence-level speech encoding without language-parallel data}.
\newblock In \emph{2024 IEEE International Conference on Acoustics, Speech and Signal Processing (ICASSP)}, pages 12722--12726, Seoul, Korea. IEEE.

\bibitem[{Kim et~al.(2017)Kim, Hori, and Watanabe}]{kim2017joint}
Suyoun Kim, Takaaki Hori, and Shinji Watanabe. 2017.
\newblock \href {https://ieeexplore.ieee.org/stamp/stamp.jsp?arnumber=7953075} {Joint ctc-attention based end-to-end speech recognition using multi-task learning}.
\newblock In \emph{2017 IEEE international conference on acoustics, speech and signal processing (ICASSP)}, pages 4835--4839. IEEE.

\bibitem[{Kong et~al.(2020)Kong, Kim, and Bae}]{Kong2020_HiFiGAN}
Jungil Kong, Jaehyeon Kim, and Jaekyoung Bae. 2020.
\newblock \href {https://doi.org/10.5555/3495724.3497152} {Hifi-gan: Generative adversarial networks for efficient and high fidelity speech synthesis}.
\newblock \emph{34th Conference on Neural Information Processing Systems (NeurIPS 2020)}.

\bibitem[{Kuhl et~al.(2008)Kuhl, Conboy, Coffey-Corina, Padden, Rivera-Gaxiola, and Nelson}]{Kuhl2008_phonetics_learning}
Patricia~K. Kuhl, Barbara~T. Conboy, Sharon Coffey-Corina, Denise Padden, Maritza Rivera-Gaxiola, and Tobey Nelson. 2008.
\newblock \href {https://doi.org/10.1098/rstb.2007.2154} {Phonetic learning as a pathway to language: New data and native language magnet theory expanded (nlm-e)}.
\newblock \emph{Philosophical Transactions of the Royal Society B: Biological Sciences}, 363:979--1000.

\bibitem[{Lavie et~al.(1997)Lavie, Waibel, Levin, Finke, Gates, Gavalda, Zeppenfeld, and Zhan}]{Laive_1997}
A.~Lavie, A.~Waibel, L.~Levin, M.~Finke, D.~Gates, M.~Gavalda, T.~Zeppenfeld, and Puming Zhan. 1997.
\newblock \href {https://doi.org/10.1109/ICASSP.1997.599557} {Janus-iii: speech-to-speech translation in multiple languages}.
\newblock In \emph{1997 IEEE International Conference on Acoustics, Speech, and Signal Processing}, volume~1, pages 99--102, Munich, Germany. IEEE.

\bibitem[{Le et~al.(2024)Le, Qian, Wang, Zhou, Liu, Wang, Yousefi, Qian, Li, Zhao et~al.}]{le2024transvip}
Chenyang Le, Yao Qian, Dongmei Wang, Long Zhou, Shujie Liu, Xiaofei Wang, Midia Yousefi, Yanmin Qian, Jinyu Li, Sheng Zhao, et~al. 2024.
\newblock \href {https://www.microsoft.com/en-us/research/publication/transvip-speech-to-speech-translation-system-with-voice-and-isochrony-preservation/} {Transvip: Speech to speech translation system with voice and isochrony preservation}.
\newblock \emph{arXiv preprint arXiv:2405.17809}.

\bibitem[{Lee et~al.(2022{\natexlab{a}})Lee, Chen, Wang, Gu, Popuri, Ma, Polyak, Adi, He, Tang, Pino, Hsu, Ai, and University}]{Lee2022}
Ann Lee, Peng-Jen Chen, Changhan Wang, Jiatao Gu, Sravya Popuri, Xutai Ma, Adam Polyak, Yossi Adi, Qing He, Yun Tang, Juan Pino, Wei-Ning Hsu, Meta Ai, and Johns~Hopkins University. 2022{\natexlab{a}}.
\newblock \href {https://doi.org/10.18653/V1/2022.ACL-LONG.235} {Direct speech-to-speech translation with discrete units}.
\newblock In \emph{Proceedings of the 60th Annual Meeting of the Association for Computational Linguistics}, volume~1, pages 3327--3339, Dublin, Ireland. Association for Computational Linguistics.

\bibitem[{Lee et~al.(2022{\natexlab{b}})Lee, Gong, Duquenne, Schwenk, Chen, Wang, Popuri, Adi, Pino, Gu, and Hsu}]{lee-etal-2022-textless}
Ann Lee, Hongyu Gong, Paul-Ambroise Duquenne, Holger Schwenk, Peng-Jen Chen, Changhan Wang, Sravya Popuri, Yossi Adi, Juan Pino, Jiatao Gu, and Wei-Ning Hsu. 2022{\natexlab{b}}.
\newblock \href {https://doi.org/10.18653/v1/2022.naacl-main.63} {Textless speech-to-speech translation on real data}.
\newblock In \emph{Proceedings of the 2022 Conference of the North American Chapter of the Association for Computational Linguistics: Human Language Technologies}, pages 860--872, Seattle, United States. Association for Computational Linguistics.

\bibitem[{Lee et~al.(2022{\natexlab{c}})Lee, Mohamed, Watanabe, Sainath, Livescu, Li, Yang, and Kirchhoff}]{Lee2022_Representation_Learning_for_Speech}
Hung~Yi Lee, Abdelrahman Mohamed, Shinji Watanabe, Tara Sainath, Karen Livescu, Shang~Wen Li, Shu~Wen Yang, and Katrin Kirchhoff. 2022{\natexlab{c}}.
\newblock \href {https://doi.org/10.18653/V1/2022.NAACL-TUTORIALS.2} {Self-supervised representation learning for speech processing}.
\newblock \emph{NAACL 2022 - 2022 Conference of the North American Chapter of the Association for Computational Linguistics: Human Language Technologies, Tutorial Abstracts}, pages 8--13.

\bibitem[{Lewis et~al.(2020)Lewis, Liu, Goyal, Ghazvininejad, Mohamed, Levy, Stoyanov, and Zettlemoyer}]{lewis-etal-2020-BART}
Mike Lewis, Yinhan Liu, Naman Goyal, Marjan Ghazvininejad, Abdelrahman Mohamed, Omer Levy, Veselin Stoyanov, and Luke Zettlemoyer. 2020.
\newblock \href {https://doi.org/10.18653/v1/2020.acl-main.703} {{BART}: Denoising sequence-to-sequence pre-training for natural language generation, translation, and comprehension}.
\newblock In \emph{Proceedings of the 58th Annual Meeting of the Association for Computational Linguistics}, pages 7871--7880, Online. Association for Computational Linguistics.

\bibitem[{Li and Liang(2021)}]{li-liang-2021-prefix}
Xiang~Lisa Li and Percy Liang. 2021.
\newblock \href {https://doi.org/10.18653/v1/2021.acl-long.353} {Prefix-tuning: Optimizing continuous prompts for generation}.
\newblock In \emph{Proceedings of the 59th Annual Meeting of the Association for Computational Linguistics and the 11th International Joint Conference on Natural Language Processing}, pages 4582--4597, Online. Association for Computational Linguistics.

\bibitem[{Li et~al.(2023)Li, Jia, and Chiu}]{Li2023_TextLess_Direct}
Xinjian Li, Ye~Jia, and Chung-Cheng Chiu. 2023.
\newblock \href {https://doi.org/10.1109/ICASSP49357.2023.10096797} {Textless direct speech-to-speech translation with discrete speech representation}.
\newblock In \emph{2023 IEEE International Conference on Acoustics, Speech and Signal Processing (ICASSP)}, pages 1--5, Rhodes Island, Greece. IEEE.

\bibitem[{Liu et~al.(2019)Liu, Xiong, Zhang, He, Wu, Wang, and Zong}]{Liu2019d_KnowledgeDistillation}
Yuchen Liu, Hao Xiong, Jiajun Zhang, Zhongjun He, Hua Wu, Haifeng Wang, and Chengqing Zong. 2019.
\newblock \href {https://doi.org/10.21437/Interspeech.2019-2582} {{End-to-End Speech Translation with Knowledge Distillation}}.
\newblock In \emph{Proceedings of Interspeech 2019}, pages 1128--1132, Graz, Austria. ISCA.

\bibitem[{Ma et~al.(2019)Ma, Huang, Xiong, Zheng, Liu, Zheng, Zhang, He, Liu, Li, Wu, and Wang}]{ma-etal-2019-stacl_AL}
Mingbo Ma, Liang Huang, Hao Xiong, Renjie Zheng, Kaibo Liu, Baigong Zheng, Chuanqiang Zhang, Zhongjun He, Hairong Liu, Xing Li, Hua Wu, and Haifeng Wang. 2019.
\newblock \href {https://doi.org/10.18653/v1/P19-1289} {{STACL}: Simultaneous translation with implicit anticipation and controllable latency using prefix-to-prefix framework}.
\newblock In \emph{Proceedings of the 57th Annual Meeting of the Association for Computational Linguistics}, pages 3025--3036, Florence, Italy. Association for Computational Linguistics.

\bibitem[{Ma et~al.(2021)Ma, Gong, Liu, Lee, Tang, Chen, Hsu, Koehn, and Pino}]{ma2022directMonotonicMultiHead}
Xutai Ma, Hongyu Gong, Danni Liu, Ann Lee, Yun Tang, Peng-Jen Chen, Wei-Ning Hsu, Phillip Koehn, and Juan Pino. 2021.
\newblock \href {https://arxiv.org/abs/2110.08250} {Direct simultaneous speech-to-speech translation with variational monotonic multihead attention}.
\newblock \emph{arXiv preprint arXiv:2110.08250}.

\bibitem[{Ma et~al.(2020{\natexlab{a}})Ma, Pino, and Koehn}]{ma-etal-2020-simulmt}
Xutai Ma, Juan Pino, and Philipp Koehn. 2020{\natexlab{a}}.
\newblock {S}imul{MT} to {S}imul{ST}: Adapting simultaneous text translation to end-to-end simultaneous speech translation.
\newblock In \emph{Proceedings of the 1st Conference of the Asia-Pacific Chapter of the Association for Computational Linguistics and the 10th International Joint Conference on Natural Language Processing}, pages 582--587, Suzhou, China. Association for Computational Linguistics.

\bibitem[{Ma et~al.(2020{\natexlab{b}})Ma, Pino, Cross, Puzon, and Gu}]{Ma2019MonotonicMA}
Xutai Ma, Juan~Miguel Pino, James Cross, Liezl Puzon, and Jiatao Gu. 2020{\natexlab{b}}.
\newblock \href {https://openreview.net/forum?id=Hyg96gBKPS} {Monotonic multihead attention}.
\newblock In \emph{International Conference on Learning Representations}. OpenReview.net.

\bibitem[{Marie et~al.(2021)Marie, Fujita, and Rubino}]{marie-etal-2021-scientific}
Benjamin Marie, Atsushi Fujita, and Raphael Rubino. 2021.
\newblock \href {https://doi.org/10.18653/v1/2021.acl-long.566} {Scientific credibility of machine translation research: A meta-evaluation of 769 papers}.
\newblock In \emph{Proceedings of the 59th Annual Meeting of the Association for Computational Linguistics and the 11th International Joint Conference on Natural Language Processing}, pages 7297--7306, Online. Association for Computational Linguistics.

\bibitem[{Nachmani et~al.(2024)Nachmani, Levkovitch, Ding, Asawaroengchai, Zen, and Ramanovich}]{nachmani_2023_translatotron3}
Eliya Nachmani, Alon Levkovitch, Yifan Ding, Chulayuth Asawaroengchai, Heiga Zen, and Michelle~Tadmor Ramanovich. 2024.
\newblock \href {https://ieeexplore.ieee.org/stamp/stamp.jsp?arnumber=10448426} {Translatotron 3: Speech to speech translation with monolingual data}.
\newblock In \emph{2024 IEEE International Conference on Acoustics, Speech and Signal Processing (ICASSP)}, pages 10686--10690. IEEE.

\bibitem[{Nakamura et~al.(2006)Nakamura, Markov, Nakaiwa, Kikui, Kawai, Jitsuhiro, Zhang, Yamamoto, Sumita, and Yamamoto}]{Nakamura2006}
Satoshi Nakamura, Konstantin Markov, Hiromi Nakaiwa, Genichiro Kikui, Hisashi Kawai, Takatoshi Jitsuhiro, Jin~Song Zhang, Hirofumi Yamamoto, Eiichiro Sumita, and Seiichi Yamamoto. 2006.
\newblock \href {https://doi.org/10.1109/TSA.2005.860774} {The atr multilingual speech-to-speech translation system}.
\newblock \emph{IEEE Transactions on Audio, Speech and Language Processing}, 14:365--375.

\bibitem[{Ney(1999)}]{Ney_1999_ST}
H.~Ney. 1999.
\newblock \href {https://doi.org/10.1109/ICASSP.1999.758176} {Speech translation: coupling of recognition and translation}.
\newblock In \emph{1999 IEEE International Conference on Acoustics, Speech, and Signal Processing. Proceedings. ICASSP}, volume~1, pages 517--520, Phoenix, AZ, USA. IEEE.

\bibitem[{Oord et~al.(2018)Oord, Li, and Vinyals}]{Van_dem_ordo_RepresentationLearning}
Aaron van~den Oord, Yazhe Li, and Oriol Vinyals. 2018.
\newblock \href {https://arxiv.org/abs/1807.03748v2} {Representation learning with contrastive predictive coding}.
\newblock \emph{arXiv preprint arXiv:1807.03748}.

\bibitem[{Oord et~al.(2017)Oord, Vinyals, and Kavukcuoglu}]{Van_Oord_2017_VQ-VAE}
Aaron Van~Den Oord, Oriol Vinyals, and Koray Kavukcuoglu. 2017.
\newblock \href {https://arxiv.org/abs/1711.00937v2} {Neural discrete representation learning}.
\newblock \emph{Advances in Neural Information Processing Systems}, pages 6307--6316.

\bibitem[{Ott et~al.(2019)Ott, Edunov, Baevski, Fan, Gross, Ng, Grangier, and Auli}]{ott-etal-2019-fairseq}
Myle Ott, Sergey Edunov, Alexei Baevski, Angela Fan, Sam Gross, Nathan Ng, David Grangier, and Michael Auli. 2019.
\newblock \href {https://doi.org/10.18653/v1/N19-4009} {fairseq: A fast, extensible toolkit for sequence modeling}.
\newblock In \emph{Proceedings of the 2019 Conference of the North {A}merican Chapter of the Association for Computational Linguistics (Demonstrations)}, pages 48--53, Minneapolis, Minnesota. Association for Computational Linguistics.

\bibitem[{Ouyang et~al.(2022)Ouyang, Wu, Jiang, Almeida, Wainwright, Mishkin, Zhang, Agarwal, Slama, Ray et~al.}]{ouyang2022training_GPT}
Long Ouyang, Jeffrey Wu, Xu~Jiang, Diogo Almeida, Carroll Wainwright, Pamela Mishkin, Chong Zhang, Sandhini Agarwal, Katarina Slama, Alex Ray, et~al. 2022.
\newblock \href {https://proceedings.neurips.cc/paper_files/paper/2022/file/b1efde53be364a73914f58805a001731-Paper-Conference.pdf} {Training language models to follow instructions with human feedback}.
\newblock \emph{Advances in neural information processing systems}, 35:27730--27744.

\bibitem[{Pan and Yang(2010)}]{Pan2010_TransferLearning}
Sinno~Jialin Pan and Qiang Yang. 2010.
\newblock \href {https://doi.org/10.1109/TKDE.2009.191} {A survey on transfer learning}.
\newblock \emph{IEEE Transactions on Knowledge and Data Engineering}, 22(10):1345--1359.

\bibitem[{Papineni et~al.(2002)Papineni, Roukos, Ward, and Zhu}]{papineni-etal-2002-bleu}
Kishore Papineni, Salim Roukos, Todd Ward, and Wei-Jing Zhu. 2002.
\newblock \href {https://doi.org/10.3115/1073083.1073135} {{B}leu: a method for automatic evaluation of machine translation}.
\newblock In \emph{Proceedings of the 40th Annual Meeting of the Association for Computational Linguistics}, pages 311--318, Philadelphia, Pennsylvania, USA. Association for Computational Linguistics.

\bibitem[{Peng et~al.(2024)Peng, Kulikov, Yang, Popuri, Lu, Wang, and Gong}]{peng2024mslms2st}
Yifan Peng, Ilia Kulikov, Yilin Yang, Sravya Popuri, Hui Lu, Changhan Wang, and Hongyu Gong. 2024.
\newblock \href {https://arxiv.org/pdf/2403.12408} {Mslm-s2st: A multitask speech language model for textless speech-to-speech translation with speaker style preservation}.
\newblock \emph{arXiv preprint arXiv:2403.12408}.

\bibitem[{Pino et~al.(2020)Pino, Xu, Ma, Dousti, and Tang}]{Pino2020_ST_4ST}
Juan Pino, Qiantong Xu, Xutai Ma, Mohammad~Javad Dousti, and Yun Tang. 2020.
\newblock \href {https://doi.org/10.21437/Interspeech.2020-2938} {{Self-Training for End-to-End Speech Translation}}.
\newblock In \emph{Proceedings of Interspeech 2020}, pages 1476--1480, Shanghai, China. ISCA.

\bibitem[{Popović(2015)}]{Popovic_2015_chrF}
Maja Popović. 2015.
\newblock \href {https://doi.org/10.18653/V1/W15-3049} {chrf: character n-gram f-score for automatic mt evaluation}.
\newblock \emph{10th Workshop on Statistical Machine Translation, WMT 2015 at the 2015 Conference on Empirical Methods in Natural Language Processing, EMNLP 2015 - Proceedings}, pages 392--395.

\bibitem[{Popuri et~al.(2022)Popuri, Chen, Wang, Pino, Adi, Gu, Hsu, and Lee}]{Popuri2022_Enahancing_SelfSupervised}
Sravya Popuri, Peng-Jen Chen, Changhan Wang, Juan Pino, Yossi Adi, Jiatao Gu, Wei-Ning Hsu, and Ann Lee. 2022.
\newblock \href {https://doi.org/10.21437/Interspeech.2022-11032} {{Enhanced Direct Speech-to-Speech Translation Using Self-supervised Pre-training and Data Augmentation}}.
\newblock In \emph{Proc. Interspeech 2022}, pages 5195--5199, Incheon, Korea. ISCA.

\bibitem[{Post(2018)}]{Post2018_scareBLEU}
Matt Post. 2018.
\newblock \href {https://doi.org/10.18653/v1/W18-6319} {A call for clarity in reporting {BLEU} scores}.
\newblock In \emph{Proceedings of the Third Conference on Machine Translation: Research Papers}, pages 186--191, Brussels, Belgium. Association for Computational Linguistics.

\bibitem[{Potapczyk and Przybysz(2020)}]{Potapczyk2020SRPOLsSF}
Tomasz Potapczyk and Pawe{\l} Przybysz. 2020.
\newblock \href {https://aclanthology.org/2020.iwslt-1.9/} {Srpol’s system for the iwslt 2020 end-to-end speech translation task}.
\newblock In \emph{Proceedings of the 17th International Conference on Spoken Language Translation}, pages 89--94. Association for Computational Linguistics.

\bibitem[{Prabhavalkar et~al.(2024)Prabhavalkar, Hori, Sainath, Schl{\"u}ter, and Watanabe}]{Prabhavalkar2023_ASR_Survey}
Rohit Prabhavalkar, Takaaki Hori, Tara~N Sainath, Ralf Schl{\"u}ter, and Shinji Watanabe. 2024.
\newblock \href {https://ieeexplore.ieee.org/stamp/stamp.jsp?arnumber=10301513} {End-to-end speech recognition: A survey}.
\newblock \emph{IEEE/ACM Transactions on Audio, Speech, and Language Processing}, 32.

\bibitem[{Radford et~al.(2018)Radford, Narasimhan, Salimans, Sutskever et~al.}]{Openai_2018_GPT}
Alec Radford, Karthik Narasimhan, Tim Salimans, Ilya Sutskever, et~al. 2018.
\newblock \href {https://cdn.openai.com/research-covers/language-unsupervised/language_understanding_paper.pdf} {Improving language understanding by generative pre-training}.
\newblock \emph{OpenAI}.

\bibitem[{Rei et~al.(2022)Rei, de~Souza, Alves, Zerva, Farinha, Glushkova, Lavie, Coheur, and Martins}]{Rei2022_COMET-22}
Ricardo Rei, José~G.C. de~Souza, Duarte~M. Alves, Chrysoula Zerva, Ana~C. Farinha, Taisiya Glushkova, Alon Lavie, Luisa Coheur, and André~F.T. Martins. 2022.
\newblock \href {https://aclanthology.org/2022.wmt-1.52} {Comet-22: Unbabel-ist 2022 submission for the metrics shared task}.
\newblock In \emph{Conference on Machine Translation-Proceedings}, pages 578--585, Abu Dhabi, United Arab Emirates. Association for Computational Linguistics.

\bibitem[{Rei et~al.(2020)Rei, Stewart, Farinha, and Lavie}]{Rei2020_COMET}
Ricardo Rei, Craig Stewart, Ana~C. Farinha, and Alon Lavie. 2020.
\newblock \href {https://doi.org/10.18653/V1/2020.EMNLP-MAIN.213} {Comet: A neural framework for mt evaluation}.
\newblock pages 2685--2702. Association for Computational Linguistics.

\bibitem[{Ren et~al.(2020{\natexlab{a}})Ren, Hu, Tan, Qin, Zhao, Zhao, and Liu}]{Ren2020_FastSpeech2}
Yi~Ren, Chenxu Hu, Xu~Tan, Tao Qin, Sheng Zhao, Zhou Zhao, and Tie-Yan Liu. 2020{\natexlab{a}}.
\newblock \href {https://arxiv.org/abs/2006.04558v8} {Fastspeech 2: Fast and high-quality end-to-end text to speech}.
\newblock In \emph{Proceedings of International Conference on Learning Representation (ICLR-2021)}. OpenReview.net.

\bibitem[{Ren et~al.(2020{\natexlab{b}})Ren, Liu, Tan, Zhang, Qin, Zhao, and Liu}]{Ren2020_SimulSpeech}
Yi~Ren, Jinglin Liu, Xu~Tan, Chen Zhang, Tao Qin, Zhou Zhao, and Tie-Yan Liu. 2020{\natexlab{b}}.
\newblock \href {https://doi.org/10.18653/V1/2020.ACL-MAIN.350} {Simulspeech: End-to-end simultaneous speech to text translation}.
\newblock In \emph{Proceedings of the 58th Annual Meeting of the Association for Computational Linguistics}, pages 3787--3796. Association for Computational Linguistics.

\bibitem[{Ren et~al.(2019)Ren, Ruan, Tan, Qin, Zhao, Zhao, and Liu}]{Ren2019_FastSpeech}
Yi~Ren, Yangjun Ruan, Xu~Tan, Tao Qin, Sheng Zhao, Zhou Zhao, and Tie~Yan Liu. 2019.
\newblock \href {https://proceedings.neurips.cc/paper_files/paper/2019/file/f63f65b503e22cb970527f23c9ad7db1-Paper.pdf} {Fastspeech: Fast, robust and controllable text to speech}.
\newblock In \emph{33rd Conference on Neural Information Processing Systems (NeurIPS 2019)}, volume~32, Vancouver, Canada. ISCA.

\bibitem[{Resnik(1998)}]{resnik-1998-parallel}
Philip Resnik. 1998.
\newblock \href {https://link.springer.com/chapter/10.1007/3-540-49478-2_7} {Parallel strands: a preliminary investigation into mining the web for bilingual text}.
\newblock In \emph{Proceedings of the Third Conference of the Association for Machine Translation in the Americas: Technical Papers}, pages 72--82, Langhorne, PA, USA. Springer.

\bibitem[{Rix et~al.(2001)Rix, Beerends, Hollier, and Hekstra}]{rix2001perceptual}
A.W. Rix, J.G. Beerends, M.P. Hollier, and A.P. Hekstra. 2001.
\newblock \href {https://ieeexplore.ieee.org/abstract/document/941023} {Perceptual evaluation of speech quality (pesq)-a new method for speech quality assessment of telephone networks and codecs}.
\newblock In \emph{Proceedings of the 2001 IEEE International Conference on Acoustics, Speech, and Signal Processing (ICASSP)}, volume~2, pages 749--752, Salt Lake City, UT, USA. IEEE.

\bibitem[{Salesky et~al.(2021)Salesky, Mader, and Klinger}]{Salesky2021}
Elizabeth Salesky, Julian Mader, and Severin Klinger. 2021.
\newblock \href {https://doi.org/10.1109/ASRU51503.2021.9688073} {Assessing evaluation metrics for speech-to-speech translation}.
\newblock In \emph{2021 IEEE Automatic Speech Recognition and Understanding Workshop(ASRU)}, pages 733--740, Cartagena, Colombia. IEEE.

\bibitem[{Schneider et~al.(2019)Schneider, Baevski, Collobert, and Auli}]{schneider2019wav2vec}
Steffen Schneider, Alexei Baevski, Ronan Collobert, and Michael Auli. 2019.
\newblock \href {https://doi.org/10.21437/Interspeech.2019-1873} {{wav2vec: Unsupervised Pre-Training for Speech Recognition}}.
\newblock In \emph{Proceedings of Interspeech 2019}, pages 3465--3469, Graz, Austria. ISCA.

\bibitem[{Schwenk(2018)}]{schwenk-2018-filtering}
Holger Schwenk. 2018.
\newblock \href {https://doi.org/10.18653/v1/P18-2037} {Filtering and mining parallel data in a joint multilingual space}.
\newblock In \emph{Proceedings of the 56th Annual Meeting of the Association for Computational Linguistics}, volume~2, pages 228--234, Melbourne, Australia. Association for Computational Linguistics.

\bibitem[{Sellam et~al.(2020)Sellam, Das, and Parikh}]{Sellam2020_BLEURT}
Thibault Sellam, Dipanjan Das, and Ankur~P. Parikh. 2020.
\newblock \href {https://doi.org/10.18653/V1/2020.ACL-MAIN.704} {Bleurt: Learning robust metrics for text generation}.
\newblock In \emph{Proceedings of the Annual Meeting of the Association for Computational Linguistics}, pages 7881--7892. Association for Computational Linguistics.

\bibitem[{Sennrich et~al.(2016)Sennrich, Haddow, and Birch}]{sennrich-etal-2016-improving}
Rico Sennrich, Barry Haddow, and Alexandra Birch. 2016.
\newblock \href {https://doi.org/10.18653/v1/P16-1009} {Improving neural machine translation models with monolingual data}.
\newblock In \emph{Proceedings of the 54th Annual Meeting of the Association for Computational Linguistics}, pages 86--96, Berlin, Germany. Association for Computational Linguistics.

\bibitem[{Shen et~al.(2020)Shen, Jia, Chrzanowski, Zhang, Elias, Zen, and Wu}]{Shen_2020_RobustControlled_TTS}
Jonathan Shen, Ye~Jia, Mike Chrzanowski, Yu~Zhang, Isaac Elias, Heiga Zen, and Yonghui Wu. 2020.
\newblock \href {http://arxiv.org/abs/2010.04301} {Non-attentive tacotron: Robust and controllable neural {TTS} synthesis including unsupervised duration modeling}.
\newblock \emph{CoRR}, abs/2010.04301.

\bibitem[{Shen et~al.(2021)Shen, Jia, Chrzanowski, Zhang, Elias, Zen, and Wu}]{shen2021nonattentive}
Jonathan Shen, Ye~Jia, Mike Chrzanowski, Yu~Zhang, Isaac Elias, Heiga Zen, and Yonghui Wu. 2021.
\newblock \href {http://arxiv.org/abs/2010.04301} {Non-attentive tacotron: Robust and controllable neural tts synthesis including unsupervised duration modeling}.
\newblock \emph{arXiv preprint arXiv:2010.04301}.

\bibitem[{Shi et~al.(2023)Shi, Tang, Lee, Inaguma, Wang, Pino, and Watanabe}]{Shi_2023_Multiple_TTS}
Jiatong Shi, Yun Tang, Ann Lee, Hirofumi Inaguma, Changhan Wang, Juan Pino, and Shinji Watanabe. 2023.
\newblock \href {https://doi.org/10.1109/ICASSP49357.2023.10095973} {Enhancing speech-to-speech translation with multiple tts targets}.
\newblock In \emph{2023 IEEE International Conference on Acoustics, Speech and Signal Processing (ICASSP)}, pages 1--5, Rhodes Island, Greece. IEEE.

\bibitem[{Shimizu et~al.(2014)Shimizu, Neubig, Sakti, Toda, and Nakamura}]{shimizu-etal-2014-collection}
Hiroaki Shimizu, Graham Neubig, Sakriani Sakti, Tomoki Toda, and Satoshi Nakamura. 2014.
\newblock \href {http://www.lrec-conf.org/proceedings/lrec2014/pdf/162_Paper.pdf} {Collection of a simultaneous translation corpus for comparative analysis}.
\newblock In \emph{Proceedings of the Ninth International Conference on Language Resources and Evaluation ({LREC}'14)}, pages 670--673, Reykjavik, Iceland. European Language Resources Association (ELRA).

\bibitem[{Sohn et~al.(1999)Sohn, Kim, and Sung}]{Sohn1999ASM}
Jongseo Sohn, Nam~Soo Kim, and Wonyong Sung. 1999.
\newblock \href {https://ieeexplore.ieee.org/abstract/document/736233} {A statistical model-based voice activity detection}.
\newblock \emph{IEEE Signal Processing Letters}, 6:1--3.

\bibitem[{Song et~al.(2023)Song, Ren, Lei, Wang, Wei, Xie, Yin, and Ma}]{song_2023_styles2st}
Kun Song, Yi~Ren, Yi~Lei, Chunfeng Wang, Kun Wei, Lei Xie, Xiang Yin, and Zejun Ma. 2023.
\newblock \href {https://doi.org/10.21437/INTERSPEECH.2023-648} {Styles2st: Zero-shot style transfer for direct speech-to-speech translation}.
\newblock In \emph{proceedings of INTERSPEECH 2023}, pages 42--46. ISCA.

\bibitem[{Taal et~al.(2011)Taal, Hendriks, Heusdens, and Jensen}]{taal2011algorithm}
Cees~H. Taal, Richard~C. Hendriks, Richard Heusdens, and Jesper Jensen. 2011.
\newblock An algorithm for intelligibility prediction of time–frequency weighted noisy speech.
\newblock \emph{IEEE Transactions on Audio, Speech, and Language Processing}, 19(7):2125--2136.

\bibitem[{Tampuu et~al.(2020)Tampuu, Semikin, Muhammad, Fishman, and Matiisen}]{Tampuu2020_End-to-End_Driving}
Ardi Tampuu, Maksym Semikin, Naveed Muhammad, Dmytro Fishman, and Tambet Matiisen. 2020.
\newblock \href {https://doi.org/10.1109/TNNLS.2020.3043505} {{A Survey of End-to-End Driving: Architectures and Training Methods}}.
\newblock \emph{IEEE Transactions on Neural Networks and Learning Systems}, 33(4):1364--1384.

\bibitem[{Tan and Koehn(2022)}]{tan2022bitext}
Weiting Tan and Philipp Koehn. 2022.
\newblock \href {https://arxiv.org/pdf/2208.11194} {Bitext mining for low-resource languages via contrastive learning}.
\newblock \emph{arXiv preprint arXiv:2208.11194}.

\bibitem[{Tang et~al.(2022)Tang, Gong, Dong, Wang, Hsu, Gu, Baevski, Li, Mohamed, Auli, and Pino}]{tang-etal-2022-unified}
Yun Tang, Hongyu Gong, Ning Dong, Changhan Wang, Wei-Ning Hsu, Jiatao Gu, Alexei Baevski, Xian Li, Abdelrahman Mohamed, Michael Auli, and Juan Pino. 2022.
\newblock \href {https://doi.org/10.18653/v1/2022.acl-long.105} {Unified speech-text pre-training for speech translation and recognition}.
\newblock In \emph{Proceedings of the 60th Annual Meeting of the Association for Computational Linguistics}, pages 1488--1499, Dublin, Ireland. Association for Computational Linguistics.

\bibitem[{Tjandra et~al.(2019)Tjandra, Sakti, and Nakamura}]{Tjandra_2019_untranscribe}
Andros Tjandra, Sakriani Sakti, and Satoshi Nakamura. 2019.
\newblock \href {https://doi.org/10.1109/ASRU46091.2019.9003853} {Speech-to-speech translation between untranscribed unknown languages}.
\newblock In \emph{2019 IEEE Automatic Speech Recognition and Understanding Workshop (ASRU)}, pages 593--600, Singapore. IEEE.

\bibitem[{Treviso et~al.(2023)Treviso, Lee, Ji, van Aken, Cao, Ciosici, Hassid, Heafield, Hooker, Raffel, Martins, Martins, Forde, Milder, Simpson, Slonim, Dodge, Strubell, Balasubramanian, Derczynski, Gurevych, and Schwartz}]{Treviso2023_NLP_Methods_Survey}
Marcos Treviso, Ji-Ung Lee, Tianchu Ji, Betty van Aken, Qingqing Cao, Manuel~R. Ciosici, Michael Hassid, Kenneth Heafield, Sara Hooker, Colin Raffel, Pedro~H. Martins, André F.~T. Martins, Jessica~Zosa Forde, Peter Milder, Edwin Simpson, Noam Slonim, Jesse Dodge, Emma Strubell, Niranjan Balasubramanian, Leon Derczynski, Iryna Gurevych, and Roy Schwartz. 2023.
\newblock \href {https://doi.org/10.1162/TACL_A_00577} {Efficient methods for natural language processing: A survey}.
\newblock \emph{Transactions of the Association for Computational Linguistics}, 11:826--860.

\bibitem[{Tsiamas et~al.(2022{\natexlab{a}})Tsiamas, G{\'a}llego, Escolano, Fonollosa, and Costa-juss{\`a}}]{tsiamas-etal-2022-pretrained}
Ioannis Tsiamas, Gerard~I. G{\'a}llego, Carlos Escolano, Jos{\'e} Fonollosa, and Marta~R. Costa-juss{\`a}. 2022{\natexlab{a}}.
\newblock \href {https://doi.org/10.18653/v1/2022.iwslt-1.23} {Pretrained speech encoders and efficient fine-tuning methods for speech translation: {UPC} at {IWSLT} 2022}.
\newblock In \emph{Proceedings of the 19th International Conference on Spoken Language Translation (IWSLT 2022)}, pages 265--276, Dublin, Ireland. Association for Computational Linguistics.

\bibitem[{Tsiamas et~al.(2022{\natexlab{b}})Tsiamas, Gállego, Fonollosa, and Costa-jussà}]{Tsiamas2022SHASAO}
Ioannis Tsiamas, Gerard~I. Gállego, José A.~R. Fonollosa, and Marta~R. Costa-jussà. 2022{\natexlab{b}}.
\newblock \href {https://doi.org/10.21437/Interspeech.2022-59} {{SHAS: Approaching optimal Segmentation for End-to-End Speech Translation}}.
\newblock In \emph{Proceedings of Interspeech 2022}, pages 106--110, Incheon, Korea. ISCA.

\bibitem[{Vaswani et~al.(2017)Vaswani, Shazeer, Parmar, Uszkoreit, Jones, Gomez, Kaiser, and Polosukhin}]{Vaswani2017_Attention}
Ashish Vaswani, Noam Shazeer, Niki Parmar, Jakob Uszkoreit, Llion Jones, Aidan~N Gomez, \L~ukasz Kaiser, and Illia Polosukhin. 2017.
\newblock \href {https://proceedings.neurips.cc/paper_files/paper/2017/file/3f5ee243547dee91fbd053c1c4a845aa-Paper.pdf} {Attention is all you need}.
\newblock In \emph{Advances in Neural Information Processing Systems}, volume~30, Long Beach, CA, USA. Curran Associates, Inc.

\bibitem[{Wahlster(2013)}]{Wahlster2000_Verbmobli}
Wolfgang Wahlster. 2013.
\newblock \emph{Verbmobil: foundations of speech-to-speech translation}.
\newblock Springer Science \& Business Media.

\bibitem[{Wang et~al.(2023{\natexlab{a}})Wang, Inaguma, Chen, Kulikov, Tang, Hsu, Auli, and Pino}]{Wang2022SimpleAE}
Changhan Wang, Hirofumi Inaguma, Peng-Jen Chen, Ilia Kulikov, Yun Tang, Wei-Ning Hsu, Michael Auli, and Juan Pino. 2023{\natexlab{a}}.
\newblock \href {https://doi.org/10.18653/v1/2023.acl-long.602} {Simple and effective unsupervised speech translation}.
\newblock In \emph{Proceedings of the 61st Annual Meeting of the Association for Computational Linguistics}, pages 10771--10784, Toronto, Canada. Association for Computational Linguistics.

\bibitem[{Wang et~al.(2021{\natexlab{a}})Wang, Riviere, Lee, Wu, Talnikar, Haziza, Williamson, Pino, and Dupoux}]{wang-etal-2021-voxpopuli}
Changhan Wang, Morgane Riviere, Ann Lee, Anne Wu, Chaitanya Talnikar, Daniel Haziza, Mary Williamson, Juan Pino, and Emmanuel Dupoux. 2021{\natexlab{a}}.
\newblock \href {https://doi.org/10.18653/v1/2021.acl-long.80} {{V}ox{P}opuli: A large-scale multilingual speech corpus for representation learning, semi-supervised learning and interpretation}.
\newblock In \emph{Proceedings of the 59th Annual Meeting of the Association for Computational Linguistics and the 11th International Joint Conference on Natural Language Processing (Volume 1: Long Papers)}, pages 993--1003, Online. Association for Computational Linguistics.

\bibitem[{Wang et~al.(2021{\natexlab{b}})Wang, Wu, Gu, and Pino}]{Wang2020CoVoST2A}
Changhan Wang, Anne Wu, Jiatao Gu, and Juan Pino. 2021{\natexlab{b}}.
\newblock \href {https://doi.org/10.21437/Interspeech.2021-2027} {{CoVoST 2 and Massively Multilingual Speech Translation}}.
\newblock In \emph{Proceedings of Interspeech 2021}, pages 2247--2251, Brno, Czechia. ISCA.

\bibitem[{Wang et~al.(2023{\natexlab{b}})Wang, Li, Wu, Hovy, and Sun}]{Wang2022_Pre-Training_Application}
Haifeng Wang, Jiwei Li, Hua Wu, Eduard Hovy, and Yu~Sun. 2023{\natexlab{b}}.
\newblock \href {https://doi.org/https://doi.org/10.1016/j.eng.2022.04.024} {Pre-trained language models and their applications}.
\newblock \emph{Engineering}, 25:51--65.

\bibitem[{Wang et~al.(2023{\natexlab{c}})Wang, Bai, Huang, Li, Hong, and Zhao}]{wang2023speechtospeech}
Yongqi Wang, Jionghao Bai, Rongjie Huang, Ruiqi Li, Zhiqing Hong, and Zhou Zhao. 2023{\natexlab{c}}.
\newblock \href {https://arxiv.org/abs/2309.07566} {Speech-to-speech translation with discrete-unit-based style transfer}.
\newblock \emph{arXiv preprint arXiv:2309.07566}.

\bibitem[{Wei et~al.(2022)Wei, Tay, Bommasani, Raffel, Zoph, Borgeaud, Yogatama, Bosma, Zhou, Metzler, hsin Chi, Hashimoto, Vinyals, Liang, Dean, and Fedus}]{Wei2022EmergentAO}
Jason Wei, Yi~Tay, Rishi Bommasani, Colin Raffel, Barret Zoph, Sebastian Borgeaud, Dani Yogatama, Maarten Bosma, Denny Zhou, Donald Metzler, Ed~Huai hsin Chi, Tatsunori Hashimoto, Oriol Vinyals, Percy Liang, Jeff Dean, and William Fedus. 2022.
\newblock \href {https://openreview.net/pdf?id=yzkSU5zdwD} {Emergent abilities of large language models}.
\newblock \emph{Transactions on Machine Learning Research}.

\bibitem[{Wei et~al.(2023)Wei, Zhou, Zhang, Chen, Liu, He, Li, and Wei}]{Wei_2013_joint_pre-train_s2st}
Kun Wei, Long Zhou, Ziqiang Zhang, Liping Chen, Shujie Liu, Lei He, Jinyu Li, and Furu Wei. 2023.
\newblock \href {https://doi.org/10.1109/ICASSP49357.2023.10095616} {Joint pre-training with speech and bilingual text for direct speech to speech translation}.
\newblock In \emph{2023 IEEE International Conference on Acoustics, Speech and Signal Processing (ICASSP)}, pages 1--5. IEEE.

\bibitem[{Wu et~al.(2023)Wu, Chang, Wu, and Lee}]{Wu2023_SpeechGen}
Haibin Wu, Kai-Wei Chang, Yuan-Kuei Wu, and Hung-yi Lee. 2023.
\newblock \href {https://arxiv.org/abs/2306.02207v3} {Speechgen: Unlocking the generative power of speech language models with prompts}.
\newblock \emph{arXiv preprint arXiv:2306.02207}.

\bibitem[{Yallop and Fletcher(2007)}]{yallop2007_phonetics_phonology}
Colin Yallop and Janet Fletcher. 2007.
\newblock \emph{An Introduction to Phonetics and Phonology}.
\newblock Macquarie University.

\bibitem[{Yang et~al.(2022)Yang, Song, King, and Xu}]{yang2022survey_semi_super}
Xiangli Yang, Zixing Song, Irwin King, and Zenglin Xu. 2022.
\newblock \href {https://ieeexplore.ieee.org/stamp/stamp.jsp?arnumber=9941371} {A survey on deep semi-supervised learning}.
\newblock \emph{IEEE Transactions on Knowledge and Data Engineering}, 35(9):8934--8954.

\bibitem[{Zanon~Boito et~al.(2020)Zanon~Boito, Havard, Garnerin, Le~Ferrand, and Besacier}]{zanon-boito-etal-2020-mass}
Marcely Zanon~Boito, William Havard, Mahault Garnerin, {\'E}ric Le~Ferrand, and Laurent Besacier. 2020.
\newblock \href {https://aclanthology.org/2020.lrec-1.799} {{M}a{SS}: A large and clean multilingual corpus of sentence-aligned spoken utterances extracted from the {B}ible}.
\newblock In \emph{Proceedings of the Twelfth Language Resources and Evaluation Conference}, pages 6486--6493, Marseille, France. European Language Resources Association.

\bibitem[{Zhang et~al.(2021)Zhang, Tan, Ren, Qin, Zhang, and Liu}]{Zhang2021_UWSpeech}
Chen Zhang, Xu~Tan, Yi~Ren, Tao Qin, Kejun Zhang, and Tie~Yan Liu. 2021.
\newblock \href {https://doi.org/10.1609/AAAI.V35I16.17684} {Uwspeech: Speech to speech translation for unwritten languages}.
\newblock In \emph{Proceedings of the AAAI Conference on Artificial Intelligence}, volume~35, pages 14319--14327. Association for the Advancement of Artificial Intelligence.

\bibitem[{Zhang et~al.(2022)Zhang, Pan, Yin, and Ma}]{zhang_2022_direct_botelneck}
Junhui Zhang, Junjie Pan, Xiang Yin, and Zejun Ma. 2022.
\newblock \href {https://arxiv.org/abs/2212.05805} {Direct speech-to-speech translation without textual annotation using bottleneck features}.
\newblock \emph{arXiv preprint arXiv:2212.05805}.

\bibitem[{Zhang et~al.(2023)Zhang, Zhou, Wang, Chen, Wu, Liu, Chen, Liu, Wang, Li et~al.}]{zhang2023speak_forign_Languages}
Ziqiang Zhang, Long Zhou, Chengyi Wang, Sanyuan Chen, Yu~Wu, Shujie Liu, Zhuo Chen, Yanqing Liu, Huaming Wang, Jinyu Li, et~al. 2023.
\newblock \href {https://arxiv.org/abs/2303.03926} {Speak foreign languages with your own voice: Cross-lingual neural codec language modeling}.
\newblock \emph{arXiv preprint arXiv:2303.03926}.

\bibitem[{Zheng et~al.(2020)Zheng, Ma, Zheng, Liu, Yuan, Church, and Huang}]{zheng-etal-2020-fluent}
Renjie Zheng, Mingbo Ma, Baigong Zheng, Kaibo Liu, Jiahong Yuan, Kenneth Church, and Liang Huang. 2020.
\newblock \href {https://doi.org/10.18653/v1/2020.findings-emnlp.349} {Fluent and low-latency simultaneous speech-to-speech translation with self-adaptive training}.
\newblock In \emph{Findings of the Association for Computational Linguistics: EMNLP 2020}, pages 3928--3937, Online. Association for Computational Linguistics.

\bibitem[{Zheng et~al.(2019)Zheng, Tao, Wen, and Yi}]{Zheng_2019_E2E_TTS}
Yibin Zheng, Jianhua Tao, Zhengqi Wen, and Jiangyan Yi. 2019.
\newblock \href {https://doi.org/10.1109/TASLP.2019.2935807} {Forward–backward decoding sequence for regularizing end-to-end tts}.
\newblock \emph{IEEE/ACM Transactions on Audio, Speech, and Language Processing}, 27(12):2067--2079.

\bibitem[{Zhu et~al.(2023)Zhu, Gao, Zhou, Zhongyi, and Xu}]{zhu-etal-2023-diffs2ut}
Yongxin Zhu, Zhujin Gao, Xinyuan Zhou, Ye~Zhongyi, and Linli Xu. 2023.
\newblock \href {https://doi.org/10.18653/v1/2023.emnlp-main.709} {{D}iff{S}2{UT}: A semantic preserving diffusion model for textless direct speech-to-speech translation}.
\newblock In \emph{Proceedings of the 2023 Conference on Empirical Methods in Natural Language Processing}, pages 11573--11583, Singapore. Association for Computational Linguistics.

\end{thebibliography}
\bibliographystyle{acl_natbib}
\appendix
\section{Appendix}

\begin{table*}
\small
\centering
\resizebox{\textwidth}{!}{%
    \begin{tabular}{llllccc|ccc}
    \toprule
       && &&& \multicolumn{2}{c}{\textbf{Direct}} & \multicolumn{3}{c}{\textbf{Cascade Baseline}} \\ \midrule
       
       \textbf{ID}&\textbf{Paper}  &\textbf{Dataset} & \textbf{Lang.}&\textbf{Textless}&\textbf{BLEU}& \textbf{MOS}&\textbf{BLEU}&\textbf{MOS}&\textbf{Model} \\  \midrule
       
       \multicolumn{10}{c}{\textbf{Models without external data}}\\ 

        \midrule
    
    1&\citet{Jia2019}&Fisher&Es$\rightarrow$En&\checkmark&0.6&\textcolor{gray}{\xmark}&\underline{\textbf{41.4}}&4.09&ST+TTS \\
    2&\citet{Zhang2021_UWSpeech}&Fisher&Es$\rightarrow$En&\checkmark&\textbf{9.35}&\textcolor{gray}{\xmark}&\textcolor{gray}{\xmark}&\textcolor{gray}{\xmark}&\textcolor{gray}{\xmark}\\

    \midrule
    
    3&\citet{Huang2022_TranSpeech}&CVSS&En$\rightarrow$Es&\checkmark&12.77&\textcolor{gray}{\xmark}&\underline{\textbf{32.86}}&\textcolor{gray}{\xmark}&ST+TTS \\
    4&\citet{Kim_2024_TranSentence}&CVSS&En$\rightarrow$Es&\checkmark&\textbf{18.72}&\textcolor{gray}{\xmark}&\textcolor{gray}{\xmark}&\textcolor{gray}{\xmark}&\textcolor{gray}{\xmark}\\ 

    \midrule

       \multicolumn{10}{c}{\textbf{Models with external data}}\\ \midrule
        5&\citet{Jia2019}&Fisher&Es$\rightarrow$En&\textcolor{gray}{\xmark}&30.1&3.69 &41&4.09&ST+TTS\\
        6&\citet{Jia2021}&Fisher&Es$\rightarrow$En&\textcolor{gray}{\xmark}&42.9&3.98&43.3&4.04&ST+TTS \\ 
        7&\citet{Lee2022}&Fisher&Es$\rightarrow$En&\textcolor{gray}{\xmark}&39.9&3.41&43.9&3.43&ASR+MT+TTS \\ 
    8&\citet{Dong2022_Psudo_Labeled}&Fisher&Es$\rightarrow$En&\textcolor{gray}{\xmark}&43.6&3.59&41.4&\textcolor{gray}{\xmark}&ST+TTS \\ 
      9&\citet{Shi_2023_Multiple_TTS}&Fisher&Es$\rightarrow$En&\textcolor{gray}{\xmark}&41.5&\textcolor{gray}{\xmark}&\textcolor{gray}{\xmark}&\textcolor{gray}{\xmark}&\textcolor{gray}{\xmark} \\
       10&\citet{Li2023_TextLess_Direct}&Fisher&Es$\rightarrow$En&\textcolor{gray}{\xmark}&\underline{\textbf{50.3}}&\textcolor{gray}{\xmark}&\textcolor{gray}{\xmark}&\textcolor{gray}{\xmark}& \textcolor{gray}{\xmark}\\

        \midrule
        
        11&\citet{Jia2019}&Conversational&Es$\rightarrow$En&\textcolor{gray}{\xmark}&42.7&4.08&48.7&4.32&ST+TTS \\ 
        12&\citet{Jia2021}&Conversational&Es$\rightarrow$En&\textcolor{gray}{\xmark}&\textbf{55.6}&4.2&\underline{\textbf{58.8}}&4.31&ST+TTS \\

        \midrule
        13&\citet{lee-etal-2022-textless}&CoVoST&Es$\rightarrow$En&\checkmark&16.3&\textcolor{gray}{\xmark}&14.8&\textcolor{gray}{\xmark}&ST+TTS \\
        14&\citet{inaguma_2023_unity}&CoVoST&Es$\rightarrow$En&\textcolor{gray}{\xmark}&\underline{\textbf{36.4}}&\textcolor{gray}{\xmark}&32.9&\textcolor{gray}{\xmark}&ASR+MT+TTS \\
        15&\citet{Popuri2022_Enahancing_SelfSupervised}&CoVoST&Es$\rightarrow$En&\textcolor{gray}{\xmark}&33.5&3.15&33.8&3.53&ASR+MT+TTS \\

        \midrule
            16&\citet{wang2023speechtospeech} &CVSS&Es$\rightarrow$En&\textcolor{gray}{\xmark}&\underline{\textbf{23.41}}&3.87&\textcolor{gray}{\xmark}&\textcolor{gray}{\xmark}&\textcolor{gray}{\xmark}\\
            17&\citet{nachmani_2023_translatotron3}&CVSS&Es$\rightarrow$En&\textcolor{gray}{\xmark}&14.25&&6.13(unsu.)&3.14&ASR+MT+TTS\\ 
        \midrule
      18&\citet{lee-etal-2022-textless}&EuroPerl-ST&Es$\rightarrow$En&\checkmark&18.9&3.26&19.2&3.23&ST+TTS \\
        19&\citet{Popuri2022_Enahancing_SelfSupervised}&EuroPerl-ST&Es$\rightarrow$En&\textcolor{gray}{\xmark}&28.6&3.15&29.1&3.53&ASR+MT+TTS \\ 
         20&\citet{Wei_2013_joint_pre-train_s2st}&EuroPerl-ST&Es$\rightarrow$En&\textcolor{gray}{\xmark}&24.6&\textcolor{gray}{\xmark}&\textcolor{gray}{\xmark}&\textcolor{gray}{\xmark}&\textcolor{gray}{\xmark}\\
          21&\citet{inaguma_2023_unity}&EuroPerl-ST&Es$\rightarrow$En&\textcolor{gray}{\xmark}&\textbf{33.1}&\textcolor{gray}{\xmark}&\underline{\textbf{34.2}}&\textcolor{gray}{\xmark}&ASR+MT+TTS \\
           22&\citet{zhu-etal-2023-diffs2ut}&EuroPerl-ST&Es$\rightarrow$En&\checkmark&14.8&\textcolor{gray}{\xmark}&19.2&\textcolor{gray}{\xmark}&ST+TT\\
    \midrule
    23&\citet{lee-etal-2022-textless}&EuroPerl-ST&En$\rightarrow$Es&\checkmark&22.7& 4.11&21.7&4.12&ST+TTS\\
    24&\citet{Popuri2022_Enahancing_SelfSupervised}&EuroPerl-ST&En$\rightarrow$Es&\textcolor{gray}{\xmark}&32.7&4.06&32.6&3.8&ST+TTS \\
    25&\citet{Wei_2013_joint_pre-train_s2st}&EuroPerl-ST&En$\rightarrow$Es&\textcolor{gray}{\xmark}&26.6&\textcolor{gray}{\xmark}&\textcolor{gray}{\xmark}&\textcolor{gray}{\xmark}&\textcolor{gray}{\xmark} \\
   26&\citet{inaguma_2023_unity}&EuroPerl-ST&En$\rightarrow$Es&\textcolor{gray}{\xmark}&\textbf{35.3}&\textcolor{gray}{\xmark}&\underline{\textbf{36.8}}&\textcolor{gray}{\xmark}&ASR+MT+TTS \\ 
    27&\citet{zhu-etal-2023-diffs2ut}&EuroPerl-ST&En$\rightarrow$Es&\checkmark&14.5&\textcolor{gray}{\xmark}&21.7&\textcolor{gray}{\xmark}&ST+TT\\

    \midrule
     28&\citet{lee-etal-2022-textless}&EuroPerl-ST&Fr$\rightarrow$En&\checkmark&\underline{\textbf{20.3}}&3.27&19.8&3.22&ST+TTS \\
     29&\citet{zhu-etal-2023-diffs2ut}&EuroPerl-ST&Fr$\rightarrow$En&\checkmark&15.2&\textcolor{gray}{\xmark}&19.8&\textcolor{gray}{\xmark}&ST+TT\\
     
    \midrule
    
    30&\citet{lee-etal-2022-textless}&EuroPerl-ST&En$\rightarrow$Fr&\checkmark&\underline{\textbf{18.7}}&2.87&18.5&2.44&ST+TTS \\
    31&\citet{zhu-etal-2023-diffs2ut}&EuroPerl-ST&En$\rightarrow$Fr&\checkmark&13.6&\textcolor{gray}{\xmark}&18.5&\textcolor{gray}{\xmark}&ST+TT\\
    
    \midrule

    32&\citet{Popuri2022_Enahancing_SelfSupervised}&MUST-C&En$\rightarrow$Es&\textcolor{gray}{\xmark}&32.1&4.06&34.2&3.8&ASR+MT+TTS \\
    
    33&\citet{inaguma_2023_unity}&MUST-C&En$\rightarrow$Es&\textcolor{gray}{\xmark}&\underline{\textbf{36.4}}&\textcolor{gray}{\xmark}&30.8&\textcolor{gray}{\xmark}&ASR+MT+TTS \\
\bottomrule
\end{tabular}
}
\captionsetup{}
  
  \caption{Performance comparison of offline direct and cascade S2ST models. The columns under "Direct" provide details about the direct models, while the columns under "Cascade" present the performance of the cascade baseline models. Only the direct S2ST models with at least one comparable result based on the dataset and language direction are included. Models that do not have results on the standard dataset are excluded. The BLEU scores indicate translation quality, MOS scores assess the subjective speech quality, and the ``Textless" column shows whether text data was used during training. There are two types of highlighted values: bold and bold with underlines. A bold value indicates that the model is outperforming other models within the group of direct models. A bold, underlined value indicates the winner among direct and cascade within the group.}
  \label{tab:performance_table}
\end{table*}

\begin{table*}
\small
\centering
  \resizebox{\textwidth}{!}{%
    \begin{tabular}{llllcc|cccc}
    \toprule
       && &&\multicolumn{2}{c}{\textbf{Direct}} & \multicolumn{3}{|c}{\textbf{Cascade}}&\\  \midrule
     
       ID&\textbf{Paper}  &\textbf{Dataset} & \textbf{Language Pair}&\textbf{BLEU}& \textbf{MOS}&\textbf{BLEU}&\textbf{MOS}&\textbf{Model}\\  \midrule
   
       

    \multicolumn{9}{c}{\textbf{Simultaneous models}}\\ \midrule
    34&\citet{ma2022directMonotonicMultiHead}&Fisher&Es$\rightarrow$En&34&\textcolor{gray}{\xmark}&39.7&\textcolor{gray}{\xmark}&ST+TTS\\
    35&\citet{ma2022directMonotonicMultiHead}&MuST-C&Es$\rightarrow$En&18.2&\textcolor{gray}{\xmark}&24.4&\textcolor{gray}{\xmark}&ST+TTS \\
    36&\citet{agranovich2024_simultron}&MuST-C&En$\rightarrow$Es&14.7&3.35&\textcolor{gray}{\xmark}&\textcolor{gray}{\xmark}&\textcolor{gray}{\xmark} \\
    37&\citet{agranovich2024_simultron}&Conversational&Es$\rightarrow$En&51.2&3.35&\textcolor{gray}{\xmark}&\textcolor{gray}{\xmark}&\textcolor{gray}{\xmark} \\

    \midrule
    &\multicolumn{9}{c}{\textbf{LLM-based models}}\\ \midrule
    38&\citet{Wu2023_SpeechGen}&CoVoST&En$\rightarrow$Es&15.9&\textcolor{gray}{\xmark}&\textcolor{gray}{\xmark}&\textcolor{gray}{\xmark}&\textcolor{gray}{\xmark} \\
    39&\citet{zhang2023speak_forign_Languages}&EMIME&En$\rightarrow$Zh& 34.45&3.41&\textcolor{gray}{\xmark}&\textcolor{gray}{\xmark}&\textcolor{gray}{\xmark}\\
    40&\citet{zhang2023speak_forign_Languages}&EMIME&Zh$\rightarrow$En&30.66&3.54&\textcolor{gray}{\xmark}&\textcolor{gray}{\xmark}&\textcolor{gray}{\xmark}\\

    41&\citet{Dong2023_PolyVoice}&EMIME&Zh$\rightarrow$En&29.40&4.10&\textcolor{gray}{\xmark}&\textcolor{gray}{\xmark}&\textcolor{gray}{\xmark} \\
    42&\citet{Dong2023_PolyVoice}&CVSS&En$\rightarrow$Es&18.3&3.60&\textcolor{gray}{\xmark}&\textcolor{gray}{\xmark}&\textcolor{gray}{\xmark} \\
    43&\citet{peng2024mslms2st}&In-house&Es$\rightarrow$En&24.78&\textcolor{gray}{\xmark}&\textcolor{gray}{\xmark}&\textcolor{gray}{\xmark}&\textcolor{gray}{\xmark} \\
    44&\citet{gong2024seamlessexpressivelm}&In-house&Es$\rightarrow$En&20.16&3.28&\textcolor{gray}{\xmark}&\textcolor{gray}{\xmark}&\textcolor{gray}{\xmark} \\
    \midrule
      \bottomrule
               \end{tabular}
}
  \captionsetup{}
  \caption{Performance description of direct Simul-S2ST models and LLM-based direct S2ST models. The table provides a detailed performance description and a comparison with the cascade model's performance. Some LLM-based models also show results while providing target oracle text as input to the model along with source speech. However, the table only shows results without Oracle target text as input to the model.}
  \label{tab:performance_Simul_LLM}
\end{table*}

\begin{table*}
\centering
    \small
    \begin{tabular}{lllll}
      \toprule
      \textbf{ID}&\textbf{Models} &\textbf{Architecture}& \textbf{\#Parameters} & \textbf{BLEU} \\
      \midrule
      43&ASR+MT+TTS \cite{Vaswani2017_Attention} &Cascade &97M & \textbf{32.6} \\
      44&ST+TTS \cite{Wang2020CoVoST2A} &Cascade &42M & 19.80 \\ \midrule
      45&S2UT \cite{Lee2022} w/o MTL &Direct&42.8 & 0.86 \\
      46&S2UT \cite{Lee2022} w/ MTL &Direct&88.1M & \textbf{14.60} \\
      47&TF-Translatoron \cite{Jia2019} w/o MTL &Direct &63.9M & 0.40 \\
      48&TF-Translatoron \cite{Jia2019} w/ MTL &Direct &132.9M &3.78 \\ \midrule
       49&S2UT(wav2vec-L+mBART)\cite{Popuri2022_Enahancing_SelfSupervised} LNA-D &Direct &827M &21.19 \\
       50&S2UT(wav2vec-L+mBART)\cite{Popuri2022_Enahancing_SelfSupervised} LNA-E &Direct &827M&20.09 \\
       51&S2UT(wav2vec-L+mBART)\cite{Popuri2022_Enahancing_SelfSupervised} LNA-D,E &Direct&827M&23.27 \\
       52&S2UT(wav2vec-L+mBART)\cite{Popuri2022_Enahancing_SelfSupervised} Full &Direct&827M &\textbf{23.47}\\
      \bottomrule
    \end{tabular}%
  \captionsetup{font=small}
  \caption{\label{Experiment_result} Results of \textbf{Es$\rightarrow$En} language pair on \textbf{CVSS-C} dataset. (S2UT: Speech-to-unit Translation, MTL: Multitask Learning, TF: Transformer), LNA: LayerNorm Attention Module, LNA-E: LNA, and self-attention of the encoder and full decoder are fine-tuned. LNA-D: The whole encoder and the LNA and both encoder and self-attention in the decoder are fine-tuned. LNA-D,E: LNA parameters are fine-tuned on the encoder and decoder. Full: all the parameters of the encoder and decoder are fine-tuned.}
\end{table*}
\begin{table*}
   \centering
   \resizebox{\textwidth}{!}{%
   \begin{tabular}{lllllccc}
     \toprule

     \textbf{Dataset} &\textbf{Domain\& Nature} & \textbf{Src Lang.} & \textbf{Tgt Lang.} & \textbf{\#Hours}& \textbf{\#Speakers} &\textbf{Gender} \\
     \midrule
      STC \cite{shimizu-etal-2014-collection}&TED Talks, Spontaneous &Jp, En & Jp, En & 31 & 3 & Mixed \\
     MaSS \cite{zanon-boito-etal-2020-mass}&Bible, Read &15 Lang. & 15 Lang. & 150  & 80 & Mixed \\
    VoxPopuli \cite{wang-etal-2021-voxpopuli}&EP, Spontaneous& 15 Lang. & 15 Lang. & 17.3K & 80 & Mixed \\
     CVSS-C \cite{jia-etal-2022-cvss}& CV, Read/Synthesized&21 Lang. & English & 719  &1 Canon. & F\\
     CVSS-T \cite{jia-etal-2022-cvss} & CV, Read/Synthesized&21 Lang. & English & 784  & Cloned& M/F \\
    SpeechMatrix \cite{duquenne_2022_SpeechMatrix}  &EP, Spontaneous &17 Lang.  &17 Lang.  & 418k & 50 & Mixed \\
    LibriS2S \cite{jeuris-niehues-2022-libris2s}&Librivox, Read& Geerman& English & 52.5 & 42 & Mixed\\ 
    FLEURES \cite{Conneau2022_FLEURS}&Wikipedia, Read &102 Lang. & 102 Lang& 1.4k & \xmark & Mixed \\
     \bottomrule
   \end{tabular}
   }
   \captionsetup{}
   \caption{\label{Dataset_Stats} Details of the available S2ST dataset (EP: European Parliament, CV: Common Voice). The domain indicates the source of the raw data, while the nature specifies whether the speech is spontaneous, read, or synthesized. "spontaneous" means both sides of the speech are spontaneous, whereas "Read/Synthesized" means one side is Read speech and the other is synthesized speech.}
 \end{table*}
\end{document}